\documentclass[manuscript,screen]{acmart}
\AtBeginDocument{%
  }

\setcopyright{acmlicensed}
\copyrightyear{2025}
\acmYear{2025}
\acmDOI{XXXXXXX.XXXXXXX}

\acmJournal{HEALTH}
\acmVolume{TBD}
\acmNumber{TBD}
\acmArticle{TBD}

\acmISBN{978-1-4503-XXXX-X/2018/06}

\usepackage{algorithm}
\usepackage{algpseudocode}
\usepackage{subcaption}
\usepackage{amsmath}
\usepackage{booktabs}
\usepackage{multirow}
\usepackage{ulem}
\usepackage{comment}
\usepackage{graphicx}
\usepackage{amsfonts}
\usepackage{textcomp}
\usepackage{caption}
\usepackage{colortbl}
\usepackage{xcolor}
\usepackage{makecell}
\definecolor{baseline1}{gray}{0.92} 
\definecolor{baseline2}{gray}{0.85} 
\definecolor{sslcolor}{RGB}{210,235,255} 
\usepackage{appendix}
\usepackage{array} 

\begin{document}

\title{Self-Supervised Learning and Opportunistic Inference for Continuous Monitoring of Freezing of Gait in Parkinson’s Disease}

\author{Shovito Barua Soumma}
\orcid{0009-0007-1949-9795}
\email{shovito@asu.edu}
\affiliation{%
  \institution{Arizona State University}
  \city{Phoenix}
  \state{AZ}
  \country{USA}
}

\author{Daniel S. Peterson}
\orcid{0000-0002-4639-6544}
\email{daniel.peterson1@asu.edu}
\affiliation{%
  \institution{Arizona State University}
  \city{Phoenix}
  \state{AZ}
  \country{USA}
}

\author{Shyamal H. Mehta}
\orcid{0000-0002-4639-6544}
\email{Mehta.Shyamal@mayo.edu}
\affiliation{%
  \institution{Mayo Clinic}
  \city{Scottsdale}
  \state{AZ}
  \country{USA}
}

\author{Hassan Ghasemzadeh}
\orcid{0000-0002-1844-1416}
\email{Hassan.Ghasemzadeh@asu.edu}
\affiliation{%
  \institution{Arizona State University}
  \city{Phoenix}
  \state{AZ}
  \country{USA}
}
\thanks{This work was supported in part by the National Science Foundation under grant CNS-2227002 as well as the ASU College of Health Solutions. Any opinions, findings, conclusions, or recommendations expressed in this material are those of the authors and do not necessarily reflect the views of the funding organizations.}

\renewcommand{\shortauthors}{Author et al.}

\begin{abstract}
 Parkinson’s disease (PD) significantly affects patients' quality of life through debilitating motor symptoms, such as Freezing of Gait (FoG). Continuous, in-home monitoring of FoG is essential for timely clinical intervention but remains challenging due to high power consumption, annotation cost, and the controlled environments required by current wearables. We introduce \textit{LIFT-PD}\footnote{The source code is available at: \href{https://github.com/shovito66/LIFT-PD}{\textcolor{blue}{https://github.com/shovito66/LIFT-PD}}}, a novel self-supervised learning \textcolor{black}{(SSL)} framework for real-time, patient-independent FoG detection that uniquely utilizes a single waist-worn accelerometer—an approach traditionally considered less optimal due to weaker gait signatures. LIFT-PD leverages \textcolor{black}{SSL} on unlabeled data collected from uncontrolled, real-world settings and employs a novel Differential Hopping Windowing Technique (DHWT) to address gait variability and dataset imbalance. Additionally, an opportunistic inference module selectively activates the deep learning model only during patient movement, significantly reducing power consumption and enabling continuous monitoring ($>$48 hours).
Experimental results show that LIFT-PD achieves a 7.25\% increase in precision and 4.4\% improvement in accuracy compared to supervised and semi-supervised baseline models while requiring approximately 40\% fewer labeled training samples. Evaluations across diverse patient characteristics—including severity, medication state, age, and gender—confirm the model’s robustness and clinical applicability, positioning LIFT-PD as a practical, energy-efficient, and scalable solution for continuous real-world FoG monitoring in PD.
\end{abstract}



\received{1 August 2025}
\received[revised]{X Y 2025}
\received[accepted]{X Y 2025}

\maketitle

\section{Introduction}
\label{sec:introduction}
Parkinson`s disease (PD) is a progressive neurological disorder affecting 7–10 million people worldwide, significantly impacting their quality of life~\cite{soumma2024mds}. Freezing of Gait (FoG), a transient inability to produce effective steps during walking~\cite{nieuwboer2013characterizing}, is one of the most debilitating symptoms of PD, leading to poor mobility, increased risk of falls and injuries, and reduced quality of life. While treatments like levodopa can sometimes reduce the severity of freezing episodes, their effectiveness is often incomplete and variable, diminishing over time~\cite{zhang2016freezing}. 
Compensatory treatments such as on-demand cueing require patient or companion initiation, which can be challenging in time-sensitive or anxiety-provoking situations that trigger freezing~\cite{cosentino2023one}. Identifying FoG events and, more importantly, the time leading up to a \textcolor{red}{FoG} event could result in early deployment of on-demand cues to help reduce the severity of a freezing event~\cite{jadhwani2023review}. 

In recent years, wearable technology and machine learning (ML) algorithms have emerged as promising tools for continuous monitoring and management of PD symptoms, including gait disturbances and tremors~\cite{borzi2021prediction,bikias2021deepfog,SilvadeLima2017FreezingOG}
However, real-world deployment of these technologies for continuous monitoring and intervention has remained a significant research challenge. One major hurdle is the scarcity of labeled data essential for training robust ML models, particularly in PD, where symptom manifestation is highly individualized. Data annotation requires considerable time and expertise, limiting the availability of accurate and diverse datasets~\cite{peppes2023foggan}.

Detecting FoG using a patient-independent model is a complex task due to significant inter- and intra-variability in patients' gait patterns. Previous research on FoG detection relied on multiple sensors, extensive feature engineering, patient-specific data collection, and model retraining, limiting large scale adoption of wearable technologies for long-term health monitoring~\cite{sigcha2022improvement, naghavi2021towards, mikos2017real}. To address these challenges, we present an innovative label-efficient, patient-independent, and robust self-supervised learning framework, \textbf{\textit{LIFT-PD}} (\textbf{L}abel-efficient \textbf{I}n-home \textbf{F}reezing-of-gait \textbf{T}racking in \textbf{P}arkinson's \textbf{D}isease), for detecting FoG events in real-time. 

The main contributions of this work are:
\begin{enumerate}
    \item \textbf{Label-Efficient Learning}:  LIFT-PD uses a self-supervised learning approach, allowing the model to learn from unlabeled data with minimal labeled instances required for fine-tuning. This significantly reduces the need for extensive data annotation, making it more practical for real-world deployment where labeled data is often limited.
    
    \item \textbf{Handling Variability and Imbalanced Data}:  Our Differential Hopping Windowing Technique (DHWT) effectively handles inter and intra-patient gait variations and data imbalance issues.
    
    \item \textbf{Robust Performance:} Despite using significantly fewer labeled data points, LIFT-PD achieves comparable performance to fully supervised approaches.
    
    \item \textbf{Energy Efficiency:} LIFT-PD includes an opportunistic inference module that activates the model only during active movement periods, significantly reducing computational load and power consumption, paving the way for implementation in a stand-alone wearable system for real-time symptom monitoring.
    
    \item \textbf{Practical Deployment:} By using a single triaxial accelerometer and processing data in real-time without extensive preprocessing, LIFT-PD offers a practical, scalable solution for continuous, in-home real-time monitoring of PD patients, facilitating seamless integration into everyday healthcare systems.    
\end{enumerate}

These contributions make LIFT-PD a scalable, energy-efficient, and patient-independent solution for real-time FoG detection, with significant potential for improving Parkinson’s disease management through continuous and accessible monitoring.
\begin{figure*}[!t]
\vspace{-3mm}
\centering
\includegraphics[width=1.03\linewidth] {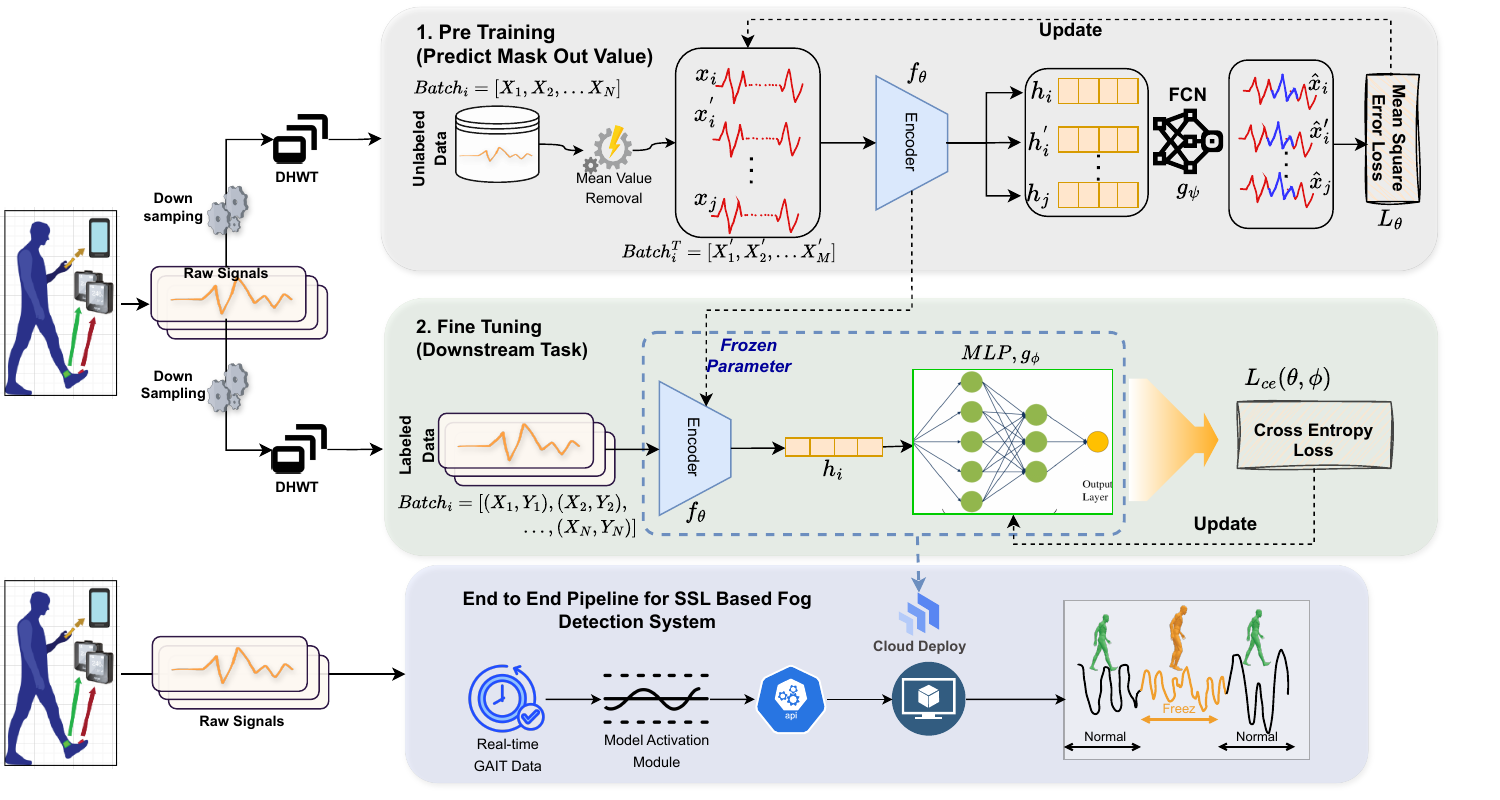}
\caption{Self-Supervised training pipeline for real-time FoG detection.
\textbf{Pretrain:} 
Encoder reconstructs masked signal segments, outputting predicted values \(h_i\) for missing data.
\textbf{Fine-tuning:} Encoder weights are frozen while the MLP is optimized using cross-entropy loss on labeled data. \textbf{End-to-End Pipeline:} The trained model is integrated into a real-time system, where \textit{model activation module (MAM)} selectively activates the computationally intensive FoG detection model for energy-efficient, long-term monitoring.}
\label{fig:method}
\vspace{-3mm}
\end{figure*}

\begin{table*}[h]
    \centering
    \caption{Summary of Recent FoG Detection Works. FI:Freezing Index, SP: Stride Peak, STD:Standard Deviation, SD: Subject Dependent, SI: Subject Independent, LOOCV: Leave-One-Out Cross-Validation}
    \setlength{\tabcolsep}{0.2pt} 
\renewcommand{\arraystretch}{1.2} 
    \begin{tabular}{cccccp{5cm}}
    
    \toprule
         \makecell{Study} & \makecell{Method} & \makecell{Sensor (Number) \\ (Position)} & \makecell{Preprocessing/\\ Extracted Features} & \makecell{Validation} & \makecell{Contribution} \\
        \midrule
\raisebox{-10pt}[0pt][0pt]{~\cite{sigcha2022improvement}} & \raisebox{-10pt}[0pt][0pt]{\makecell{CNN, Transformer \\ (Supervised)}} &   \multirow{5}{*}{\makecell{Accelerometer (1) 
\\(Left Waist)}}  & \raisebox{-10pt}[0pt][0pt]{\makecell{Resample, Filtering, \\ FFT, FI}}  & \raisebox{-10pt}[0pt][0pt]{\makecell{LOSO CV\\(SI)}} & Transformer and CNN based methodologies for FoG detection, along with a clustering approach for FoG episode analysis \\
    \raisebox{-5pt}[0pt][0pt]{~\cite{10.1016/j.artmed.2022.102459}} & \raisebox{-5pt}[0pt][0pt]{\makecell{Multihead CNN\\(Supervised)}} &   & \raisebox{-5pt}[0pt][0pt]{\makecell{ $\times$}}  & \raisebox{-5pt}[0pt][0pt]{ Hold Out (SI)} & Light algorithm for real-time FoG detection using raw sensors data. \\
         \midrule
    \raisebox{-3pt}[0pt][0pt]{~\cite{borzi2021prediction}} & \raisebox{-3pt}[0pt][0pt]{Supervised ML} & \raisebox{-3pt}[0pt][0pt]{\makecell{IMU (2)\\(Shin)}} &  \raisebox{-3pt}[0pt][0pt]{\makecell{Time \& Frequency \\Domain Features}}  & \raisebox{-3pt}[0pt][0pt]{\makecell{
    10 Fold CV, LOSO \\train-test 70-30\%}} & Detect and predict FoG considering the impact of dopaminergic therapy on performance.\\
         \midrule
         ~\cite{mikos2017real} & \raisebox{-3pt}[0pt][0pt] {\makecell{Semi Supervised\\3 Layers}} &\makecell{IMU (1)\\(Ankle)}  & \makecell{Filtering, 3 Features \\ (FI, SP, STD)}   & LOSO (SD)  & Adapting FoG classifier's parameters in real-time with unlabeled data\\
         \midrule
        ~\cite{naghavi2021towards} & \makecell{Supervised,\\transfer learning} & \makecell{IMU (2)\\(Left-Right Ankle)} &  \makecell{Filtering,\\Data Augmentation}  & \makecell{LOOCV (SI)} & {One-class classifier for FoG detection using only normal gait data}\\
         \bottomrule
    \end{tabular}
    \label{tab:pd.fog-literature}
\end{table*}
\section{Related Work}
The detection of FoG events in PD has been the subject of extensive research, often involving multimodal datasets. In particular, prior work explored the effectiveness of different sensor modalities such as gait acceleration, electroencephalogram (EEG), electromyography (EMG), and skin conductance (SC) in FoG detections~\cite{zhang2022multimodal}. Furthermore, researchers used an LSTM-based model combining acceleration and synthetic EEG data~\cite{guo2022high}. Previous work also introduced transfer learning~\cite{10331481}, data augmentation technique and resampling, which are efficient in scenarios with limited labeled data~\cite{wang2022sensor}. 
We can broadly categorize these prior works based on label usage into supervised learning approaches for motion analysis and weakly labeled approaches, including semi-supervised and self-supervised learning, for FoG detection.

Supervised methods for FoG detection often rely on large amounts of labeled data, which is a significant limitation in real-world applications. These methods typically use machine learning (ML) algorithms such as CNNs, transformers, and LSTMs~\cite{guo2022high} combined with sensor modalities like accelerometers and IMUs. Multi-head CNN~\cite{10.1016/j.artmed.2022.102459} and CNN-Transformer models~\cite{sigcha2022improvement} have shown good performance in detecting FoG in PD patients, but they face challenges in generalization and high computational requirements, which limit their feasibility for real-time wearable monitoring. While these models provide high detection rates, their reliance on extensive feature engineering~\cite{borzi2021prediction, koltermann2024gait}, a large number of training samples, and significant computational resources makes them impractical for wearable, real-time monitoring. Additionally, the models struggle with high false-positive rates, which can reduce their effectiveness in real-world settings where minimizing such errors is crucial for patient safety and usability. Transfer learning techniques, such as those used in the One-Class Classifier~\cite{naghavi2021towards}, help improve detection in normal conditions but fail to adapt to the dynamic and complex nature of FoG episodes, limiting their effectiveness in real-world applications. Table~\ref{tab:pd.fog-literature} summarizes recent PD studies on FoG detection. 

To address the challenge of insufficient labeled data, some studies have turned to semi-supervised learning (semi-SL), using a combination of labeled and unlabeled data for model training. Mikos et al.~\cite{mikos2017real} uses a three-layer network with an IMU at the ankle, incorporating filtering and feature extraction. While the approach shows improvement over purely supervised models, it still requires a considerable amount of labeled data and suffers from issues related to the inability to generalize effectively across different PD patient profile.

The application of self-supervised learning (SSL) to time-series data, particularly for human activity recognition (HAR), is a newer area of research. Unlike supervised and semi-supervised methods, SSL frameworks like SimCLR~\cite{chen2020simple} focus on learning representations without relying on labeled data, making them ideal for scenarios with limited annotations.
Originally, it was designed for computer vision tasks, but was later adapted for HAR using a transformer-based encoder method~\cite{khaertdinov2021contrastive}.  However, the use of SSL for FoG detection remains limited due to challenges in effectively balancing the dataset and ensuring model generalization to diverse populations. 
The efficacy of self-supervised learning in medical time series analysis has been demonstrated, emphasizing its role in data augmentation and contrastive pair formation~\cite{liu2023self}.


While numerous studies have proposed effective methods for FoG detection, most rely on either extensive labeled data or computationally expensive models that are unsuitable for real-time, wearable deployment. These methods often depend on data collected in controlled environments, limiting their practical utility for everyday monitoring. Additionally, existing approaches struggle with imbalanced datasets, where FoG events are underrepresented and have limited generalization across diverse patient profiles, such as varying severities, ages, and genders. In contrast, LIFT-PD introduces a self-supervised framework that addresses these challenges by leveraging a Differential Hopping Windowing Technique (DHWT) for handling class imbalances during training. By using a minimal amount of labeled data and combining Opportunistic Inference Modules to optimize power consumption, LIFT-PD achieves real-time FoG detection with significantly reduced computational overhead. LIFT-PD’s use of a single triaxial accelerometer and real-time data processing without extensive preprocessing makes it highly practical for wearable devices, unlike traditional models that rely on multiple sensors and are too resource-intensive for continuous monitoring.


\section{\textit{LIFT-PD} System Design}
\label{sec:design}


Training robust deep learning (DL) models typically requires large amounts of labeled data, which can be challenging to obtain, especially for tasks like freezing of gait (FoG) detection in elderly. To address this issue, LIFT-PD employs self-supervised learning to utilize unlabeled data during training. Another challenge that we face while designing ML models for FoG detection is that FoG events are sparse, resulting in an imbalanced dataset with a significantly lower proportion of FoG episodes compared to non-FoG activities. To mitigate this challenge, we incorporate the 'Differential Hopping Windowing Technique (DHWT) during data preprocessing, which applies variable overlaps for FoG and non-FoG instances, thereby enhancing the model's ability to learn from the underrepresented class. Finally,  an opportunistic-based lightweight algorithm is introduced to reduce execution complexity, allowing for implementation in stand-alone wearable devices (Section~\ref{sub-sec:model-activation-module}).

\vspace{-2mm}
\subsection{Problem Modeling by Self-Supervised Learning (SSL)}
\label{sub-sec:self-supervised-learning}
The FoG event detection problem is framed as a multivariate time-series classification task. At each time stamp \( t \), the input raw signal is represented as a vector
\( \mathbf{x}_t = [x_{t}^{1}, x_{t}^2, x_{t}^3]\)
where \(\mathbf{x}_t \in \mathbb{R}^{C=3} \) and 
\( \mathbf{C}\) corresponds to the three-channel (x, y, z) accelerometer data. 
These raw signals are then combined into a matrix, $\mathbf{X} = [x_1, x_2, \ldots, x_T] \in \mathbb{R}^{T \times C}$. After applying the DHWT method, the signals are transformed into \(N\) number of training frames (windows) $(\mathbf{X} \in \mathbb{R}^{T \times C} \rightarrow \mathbf{X_W} \in \mathbb{R}^{N \times T^\prime \times C})$ where \(x_{w_i} \in X_W\) represents \(i^{th}\) window, \(T^\prime\) is the windowing time length and $\mathbf{X_W} = [x_{w_1}, x_{w_2}, \ldots, x_{w_N}]$. 
\[
D: \mathbf{X} \in \mathbb{R}^{T \times C} \rightarrow \mathbf{X_W} \in \mathbb{R}^{N \times T^\prime \times C}
\]
\vspace{-5mm}
\[ \mathbf{X_W} = [x_{w_1}, x_{w_2}, \ldots, x_{w_N}]
\]The ultimate goal is to correctly assign a label \( y \in \{0,1\} \) to each window, where \( y=1 \) represents ``FoG'' and \( y=0 \) represents ``non-FoG''.

LIFT-PD uses SSL for FoG detection in two steps: learning contextual representations from raw signals using a 1D CNN model, followed by performing the downstream FoG detection task. In this paper, raw accelerometer signals are used as physiological signals and the downstream task is binary `Freezing of Gait' detection.

\textcolor{black}{
The first stage-SSL pre-training—is performed through a pretext task of masked-signal reconstruction, where the model learns meaningful temporal representations from unlabeled data.} In the following subsections, we describe in detail the two main components of our SSL approach:

\subsubsection{Pretext Task (Signal Reconstruction)}
\label{subsub-sec:pretext}
During pre-training, a 3-layer neural network model with an encoder is trained using SSL. \textcolor{black}{Random segments totaling 20\% of the window length are masked from each window (\(x_{w_i}\)) by replacing them with zeros. The model is then trained to predict these masked values as shown in Fig.~\ref{fig:method} (Part 1).} \textcolor{black}{Preliminary experiments comparing masking ratios of 10\%, 20\%, and 30\% showed that 20\% achieved the best trade-off—posing enough reconstruction difficulty to learn robust representations while preserving sufficient temporal context for meaningful feature learning.}
This whole training process does not need any labeled data. 
The masked signal data $\hat{X}_W\in\mathbb{R}^{N \times T^\prime}$, is defined by $\hat{X}_W=mask(X_W,m)$ where 
$\hat{x}_{w_i}\in\hat{X}_W$ \textcolor{black}{and $m$ represents the masking ratio (0.2 in our implementation)}. 

The encoder \( f_\theta: \mathbb{R}^{N \times T^\prime} \rightarrow \mathbb{R}^{N \times D} \) reconstructs the original signal from the masked windows, producing a lower dimensional hidden representation \(H \in \mathbb{R}^{N \times D}\),
where $h_i\in\mathbb{R}^D \leftarrow f_\theta(\hat{x}_{w_i})$. These are passed through a fully connected network \(g_\psi(\cdot)\) to predict the masked segments represented as $\hat{x} = g_\psi(h_i)$.
The encoder minimizes the reconstruction loss \(L_\theta\) using mean squared error (MSE) defined as:,
\begin{equation}
\label{eqn:finetune}
\begin{aligned}
    L_{\theta} = \frac{1}{N_m} \sum_{j=1}^{N_m} & (\hat{x}_j - x_{p(j)})^2
\end{aligned}
\end{equation}
where \(N_m\) is the total number of masked points, \(\hat{x}_j\) is the prediction for the \(j-\)th masked point, and \(x_{p(j)}\) is the original input value, whose position is \(p(j)\) in the segmented signal windows.
The objective is to find the optimal parameters \(\theta^* \) that minimize this \(L_\theta\) loss, which we can formally define as below:
\[\theta^* = \arg\min_{\theta} L_{\theta}\]

\subsubsection{Downstream Task (FoG Classification)}
\label{subsub-sec:downstream}
After pre-training, a Multilayer Perceptron (MLP) \(g_\phi : \mathbb{R}^{N \times D}\rightarrow Y \), is introduced on top of the encoder for binary FoG detection. The MLP maps the learned embeddings \(\mathbf{H}\) to FoG or non-FoG labels \(\mathbf{Y}\),
represented as $\hat{Y} = g_\phi (H) = g_\phi ( f_\theta(X_W) )$,
where \(f_\theta\) is the pre-trained encoder with parameters \(\theta\) and \(X_W\) is the segmented signal data.


This fine-tuning step is supervised, using a small amount of labeled data. Initially, the encoder \(f_\theta\) is initialized with the pre-trained weights \((\theta^*)\), while MLP \(g_\phi\) is randomly initialized. During the fine-tuning the encoder weights remains frozen and only the MLP layers are trained with a low learning rate as shown in Fig.~\ref{fig:method} (Part 2). For the downstream FoG detection task, we minimize the binary cross-entropy loss \(L_{ce}{(\theta,\phi)}\):
\begin{equation}
\label{eqn:finetune}
\begin{aligned}
    L_{ce}{(\theta,\phi)} = -\frac{1}{N}  \sum_{i=1}^{N} \left[ y_i \log(\sigma(\hat{y}_i)) + (1 - y_i) \log(1 - \sigma(\hat{y}_i)) \right]
\end{aligned}
\end{equation}
where N is the total number of windows, \(y\) is an indicator variable  (1 for FoG, 0 for non-FoG), \(\sigma\) is the sigmoid function, and \(\hat{y} \in \hat{Y}\) is the output of the classifier \(g_{\phi}(h_i)\) for the \(i\)'th window of embedding \(h_i\).

\subsection{Opportunistic Inference}
\label{sub-sec:model-activation-module}

\begin{figure}[!htb]
\vspace{-2mm}
\centering
\includegraphics[width=0.95\linewidth]{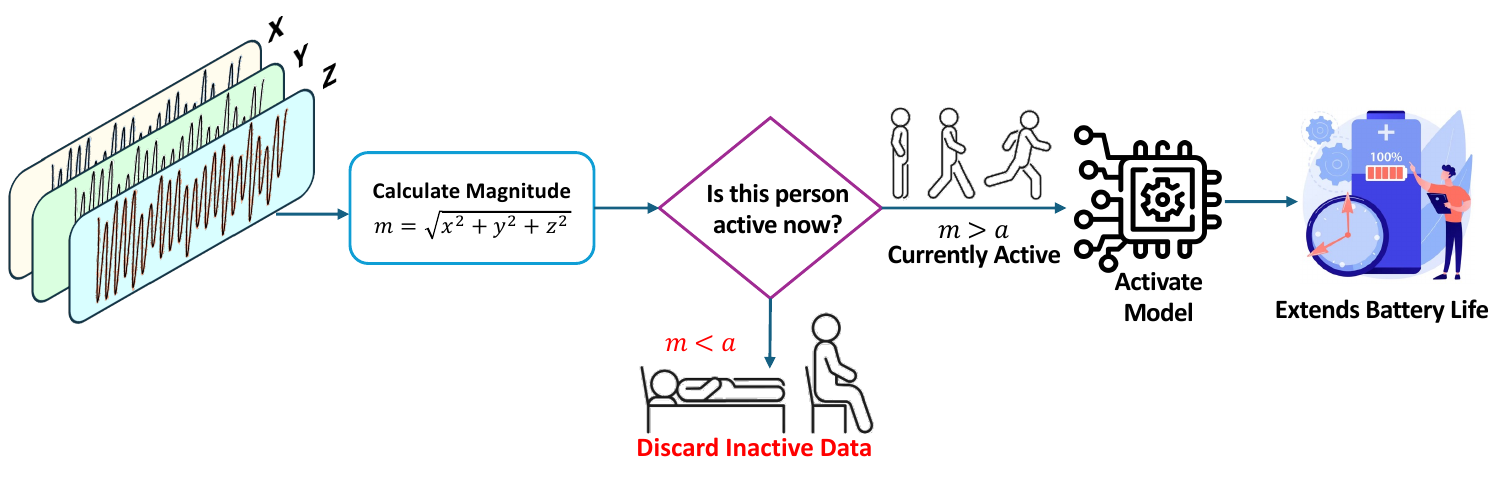}
\vspace{-2mm}
\caption{\textbf{Model Activation Module:} Activity threshold-based triggering mechanism for computational offloading and battery life extension in wearable FoG detection system.}
\label{fig:activity-threshold}
\vspace{-4mm}
\end{figure}

To optimize power consumption and computational resources for real-life in-home PD monitoring using wearable devices, we propose an opportunistic-based naive algorithm as a \textit{model activation module (MAM)}. This module differentiates between active and inactive periods, activating the computationally heavy self-supervised learning (SSL) FoG detection model only during identified active intervals, as shown in Fig.~\ref{fig:activity-threshold}.
The activation module filters out inactive windows, ensuring the SSL model is selectively executed only when activity is detected, significantly reducing power consumption and avoiding false positives during inactivity. During inactive periods, simpler methods handle the data by comparing the magnitude of the current incoming signal, leading to more efficient power utilization.
For each window \(i \in [1, N]\), the magnitude \(M_i\)  of the 3D accelerometer is calculated as $M_i = \frac{1}{T^\prime} \sum_{t=1}^{T^\prime} \left\| \sqrt{a_{x,t}^2 + a_{y,t}^2 + a_{z,t}^2} \right\|$ where \(a_x, a_y\), and \(a_z\) are the acceleration signal along each axis.
A window is considered active if \(M_i\geq\alpha\), where $\alpha$ is a predefined threshold (Eq.~\ref{eqn:conditional}), 
otherwise MAM discards it.
\begin{equation}
\label{eqn:conditional}
\begin{aligned}
\left\{
\begin{array}{ll}
1  \text{ (active)}   & \text{if } M_i \geq \alpha \\
0  \text{ (inactive)} & \text{otherwise}
\end{array}
\right.
\end{aligned}
\end{equation}
The threshold was chosen to discard windows without degrading SSL algorithm performance. Finally, the effect of the magnitude threshold was evaluated separately during the inference time (Sec.~\ref{sub-sec:result.threshold}).

\subsection{Imbalanced Training Data Mitigation}
\label{sub-sec:dhwt}




Our proposed dataset generation method (DHWT) effectively handles imbalanced datasets without the need for additional preprocessing, balancing informative features with computational efficiency for real-time deployment on wearable devices. We segment raw sensor data into short, overlapping windows, adjusting the overlap based on activity type. 
\begin{figure*}[h]
\vspace{-2mm}
\centering
    \subfloat[
    DHWT: Overlapping size increases.
    \label{subfig:dhwt}]{%
      \includegraphics[width=0.48\linewidth]{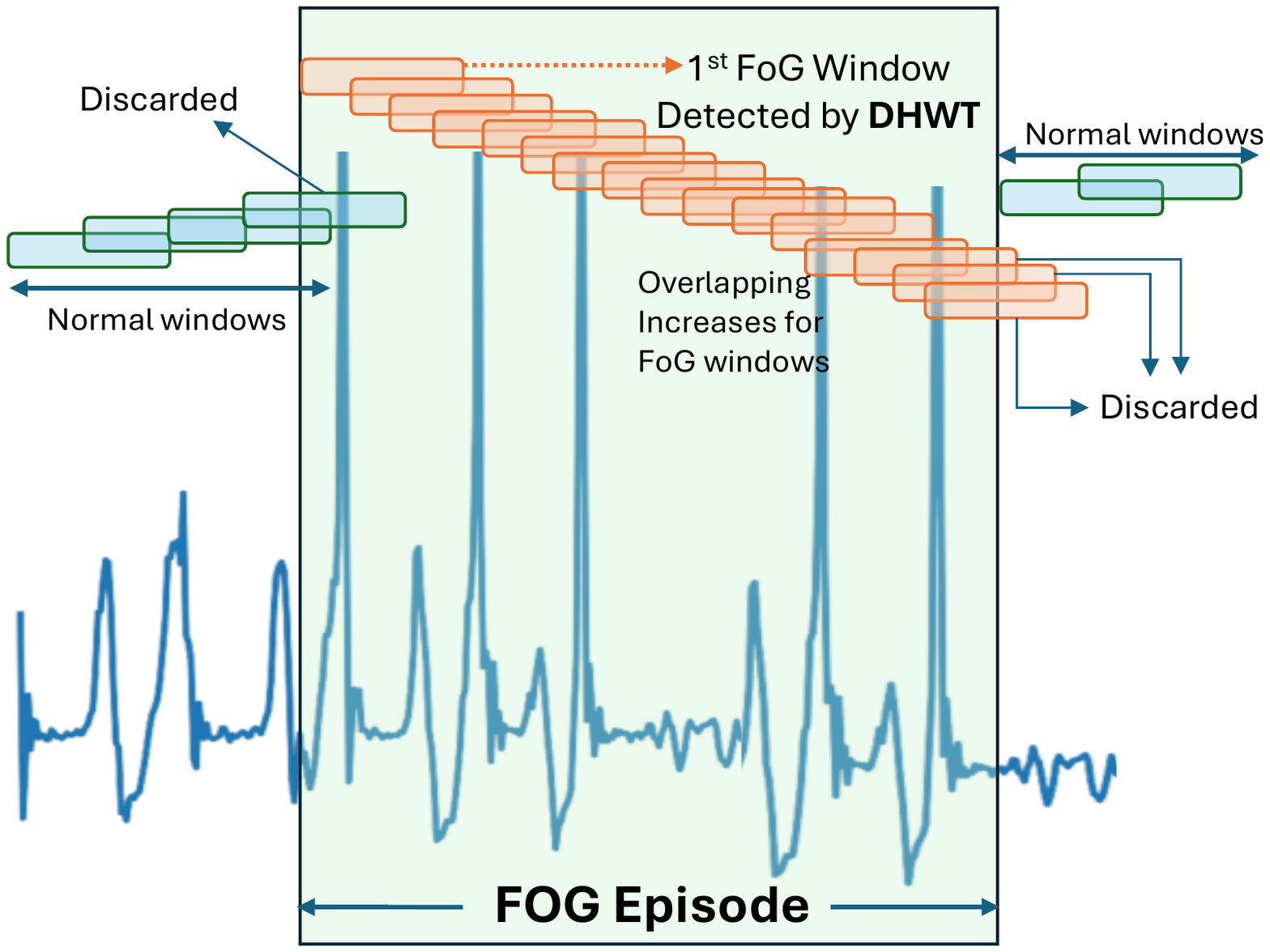}
    }
    \hfill
    \subfloat[
    Balanced class distribution using DHWT
    \label{subfig:dhw-samples}]{%
      \includegraphics[width=0.48\linewidth]{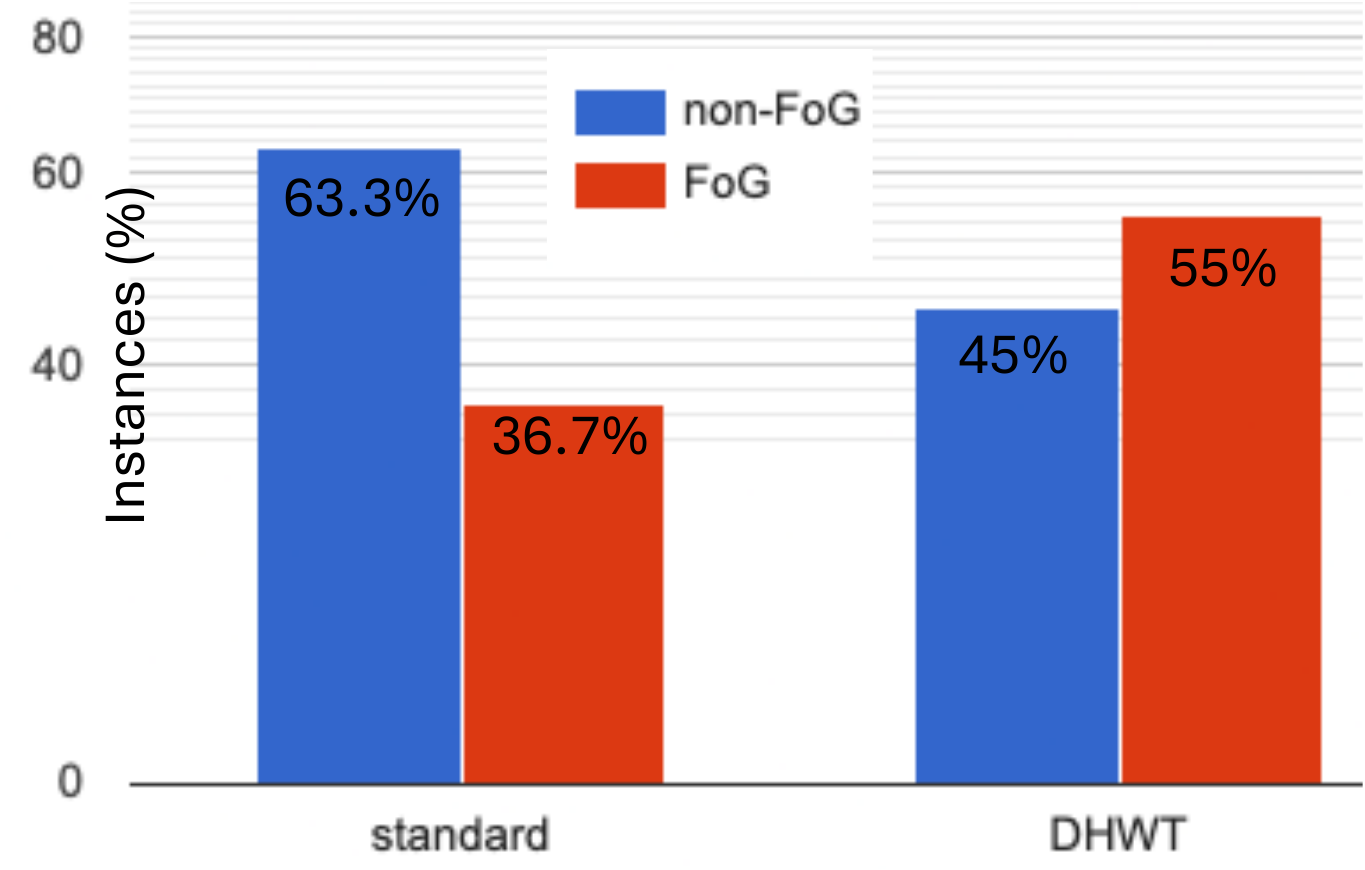}
    }
    \caption{DHWT segmentation process for training set \textbf{(a)}; FoG proportion using standard vs. DHWT segmentation \textbf{(b)}.
    }
    \label{fig:dhwt}
    \vspace{-2mm}
\end{figure*}
For instance, in our experimental evaluation shown in Fig.~\ref{subfig:dhwt}, we applied a 50\% overlap for non-FoG periods and a 75\% overlap for FoG episodes during training. During inference, windows are segmented using standard, non-variable overlaps to simulate real-world conditions where FoG episodes are not pre-identified.
Using fixed-length overlaps typically results in an unbalanced training set, as illustrated in the left bar of Fig.~\ref{subfig:dhw-samples}, with 63.3\% non-\textcolor{red}{FoG} and 36.7\% \textcolor{red}{FoG} instances. In contrast, the DHWT segmentation approach achieves a more balanced distribution, as shown in the right bar of the same figure, with non-\textcolor{red}{FoG} and \textcolor{red}{FoG} data at 45\% and 55\%, respectively. For test set generation, standard segmentation with a fixed 50\% overlap (advancing every 1.5 seconds) mimics actual operating conditions, processing data from the preceding 3 seconds. A detailed explanation of the choice of a 3-second window length is provided in Section~\ref{subsec:windowsize}.
\textcolor{blue}{
The selection of 75\% overlap for FoG and 50\% for non-FoG segments was guided by both empirical validation and domain expertise. With a 3-s window and typical gait-cycle durations of 1.2–1.5s in PD patients~\cite{yin2024gait,pistacchi2017gait}, a 75\% overlap (hop $\approx$ 0.75 s) ensures each training window contains multiple stride transitions and captures the temporal evolution of a FoG bout. In contrast, a 50\% overlap (hop $\approx$ 1.5 s) during non-FoG activity reduces sample redundancy and computation. These ratios balance class representation ($\approx$ 1:1 FoG:non-FoG windows) while maintaining physiologically meaningful temporal coverage consistent with expert-observed gait dynamics.}

\subsection{Dataset}
\label{sub-sec:dataset}

The dataset for this study is the publicly available tDCS FoG dataset, comprising movement data from PD subjects in both medicated (`On') and unmedicated (`Off') states during Freezing of Gait (FoG) provocation protocols~\cite{howard2023parkinsons}. Data were collected using a 3D accelerometer attached to the lower back, recording at 128 Hz, resulting in 1132 FoG episodes (285 minutes total) and 15.3 hours of recording. Each FoG episode was videotaped and annotated by experts~\cite{howard2023parkinsons}.
The demographic and clinical characteristics of the subjects are summarized in Table~\ref{tab:characteristic}. The dataset contains events labeled as ``Normal" or ``Freezing of Gait" (FoG). The distribution of these events is summarized in Table~\ref{tab:event-distribution}.
The sessions followed the FoG provocation protocol as described in seminal studies by Reches et al.~\cite{reches2020} and Manor et al.~\cite{manor2021multitarget}, and originally defined by Ziegler et al.~\cite{ziegler2010}.

\begin{table}[h]
\vspace{-2mm}
\caption{Summary of patient characteristics}
\label{tab:characteristic}
\centering
\setlength{\tabcolsep}{2pt} 
\renewcommand{\arraystretch}{1.2} 
    \begin{tabular}{|c|c|c|c|}
        \hline
        \raisebox{-5pt}[0pt][0pt]{\textbf{Characteristics}} & \raisebox{-5pt}[0pt][0pt]{\textbf{Overall}} & \multicolumn{2}{c|}{\textbf{Medication}} \\ \cline{3-4} 
         &  & \textbf{On} & \textbf{Off} \\ \hline
        Male (Female) & 8 (32) & 7 (31) & 6 (20) \\ \hline
        Age, mean (SD) & 69.5 (7.75) & 70.9 (6.5) & 68 (8.3) \\ \hline
        UPDRS ON, mean (SD) & 34.27 (12.7) & 34.27 (12.7) & -- \\ \hline
        UPDRS OFF, mean (STD) & 42.88 (12.99) & -- & 42.88 (12.99) \\ \hline
        NFOGQ, mean (STD) & 17.12 (7.57) & -- & -- \\ \hline
        \makecell{Years Since DX \\ mean(SD), [min, max]} & \multicolumn{3}{c|}{10.5 (5.9), [1, 23]} \\ \hline
    \end{tabular}
\vspace{-2mm}
\end{table}
\begin{table}[h]
\caption{Event distribution during FoG provocation trials}
\label{tab:event-distribution}
\centering
\setlength{\tabcolsep}{4pt} 
\renewcommand{\arraystretch}{1.2} 
    \begin{tabular}{|c|c|c|c|}
        \hline
        \raisebox{-5pt}[0pt][0pt] {\textbf{\makecell{Event Type}}} & \multicolumn{2}{c|}{\textbf{Medication (\%)}} & \raisebox{-5pt}[0pt][0pt] {\textbf{Total (\%)}} \\
\cline{2-3} 
           & \textbf{On} & \textbf{Off} & \\ \hline
        \makecell{Normal} & 28.01 & 35.41 & 63.42 \\ \hline
        \makecell{FoG} & 18.73 & 17.86 & 36.59 \\ \hline
    \end{tabular}
    \vspace{-2mm}
\end{table}

\begin{figure}[htbp]
\centering
\includegraphics[width=0.55\linewidth]{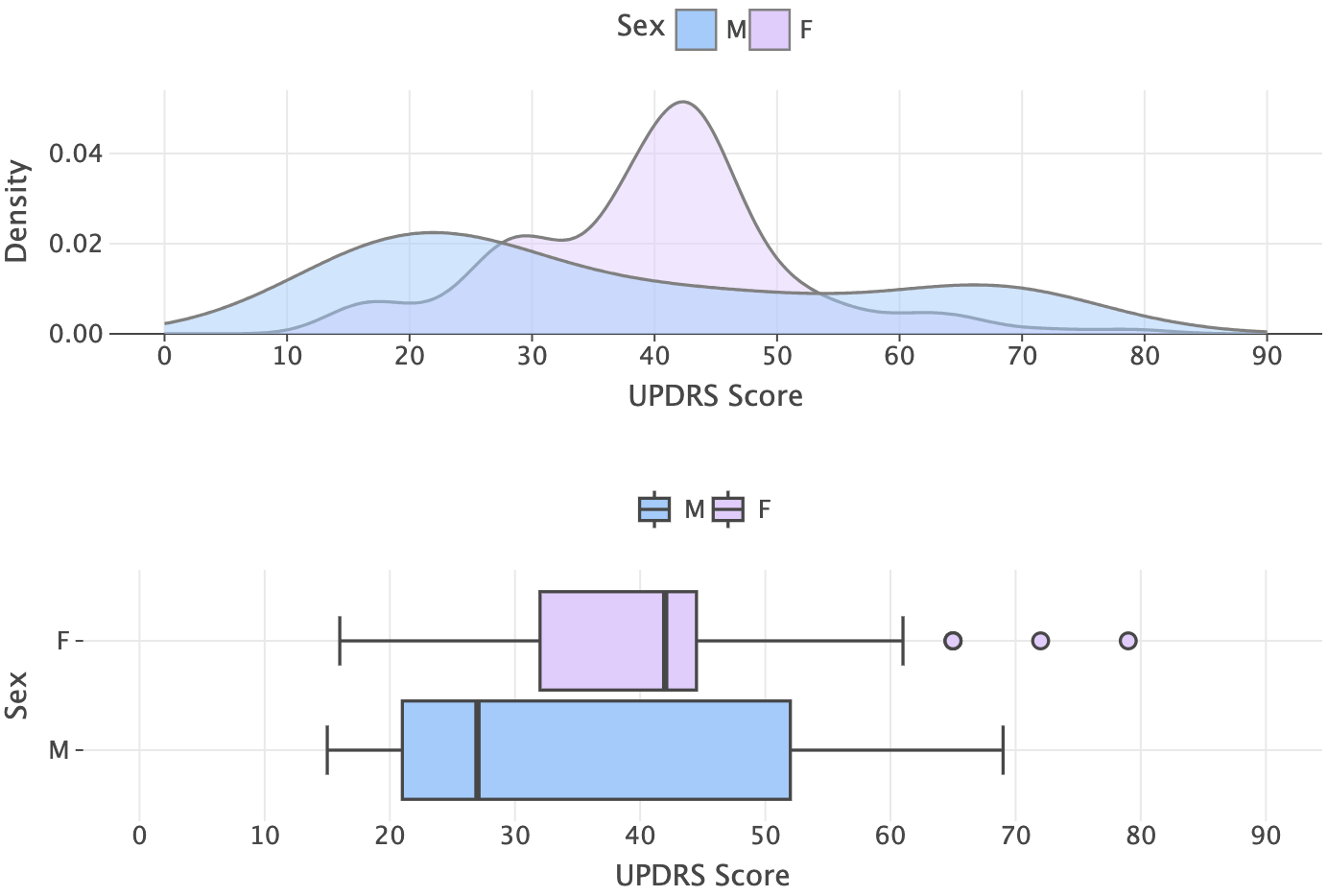}
\caption{Density distribution and box plot of the Unified Parkinson's Disease Rating Scale (UPDRS) scores, separated by sex. The top panel displays the density distribution of UPDRS scores, with males shown in blue and females in purple. The bottom panel presents a box plot indicating variations in UPDRS scores between sexes, highlighting the median scores, interquartile ranges, and outliers.}
\label{fig:sex-updrs}
\end{figure}

We resampled the data to 40 Hz, which is considered an effective frequency for recognizing human activity through accelerometers in both healthy individuals and PD patients, including those experiencing Freezing of Gait (FoG) episodes. 
This frequency captures the typical freeze band (3-8 Hz) while reducing memory load and computational complexity.
This reduced frequency is sufficient for capturing the typical freeze band (ranging from 3 to 8 Hz) that appears during FoG episodes. Additionally, downsampling to 40 Hz helps to decrease the memory load and computational complexity of the data processing.

The density plot (Fig~\ref{fig:sex-updrs}, top panel) suggests that while both distributions are roughly normal, males exhibit a higher peak density in the score range of 30-40, compared to females who show a broader peak around 40-50. The box plot (Figure~\ref{fig:sex-updrs}, bottom panel) indicates a slightly lower median UPDRS score for males compared to females, although overlaps in the interquartile ranges and the presence of outliers in both groups suggest variability within the scores.

\subsection{Mean Value Removal and Labeling}
\label{subsubsub-sec:meanvalue-label}
After segmentation, we perform minimal pre-processing on each window by removing the mean value from each of the sensor axes (e.g., x, y, z). This centering of the data mitigates sensor bias and reduces computational complexity. Finally, we assign labels to the windows based on their content: non-Fog for windows containing only non-FoG data, and FoG for windows with at least 50\% FoG data. Any windows containing a mix of both activities are discarded to ensure clear classification during the machine learning stage. This minimal preprocessing approach prioritizes real-time performance while preserving features relevant for FoG detection.

\subsection{ Leave One Group Out (LOGO)}
\label{sub-sec:logo}
In order to evaluate the subject-independence, we perform a leave-one-group-out cross-validation. The whole tdcs dataset is divided into two groups, each of them consisting of randomly chosen 20 patients. The two groups are generated so that patients in each group have similar characteristics in terms of age, duration of symptoms, and UPDRS. Hence, the patient-independent model is trained using 20 patients from each group using SSL, while the left-out group is stripped of its labels for validation. The classification performance for the left-out groups of patients is then evaluated and compared with our baseline supervised model. This procedure is repeated for each group within the dataset, conducted three times, and the results are averaged to ensure reliability and robustness in our findings.

\subsection{Model Development \& Experimental Setup}
\label{sub-sec:experimental-setup}

\begin{figure}[!htb]
    \centering
    \includegraphics[width=.95\linewidth, trim={75 128 75 108},clip]{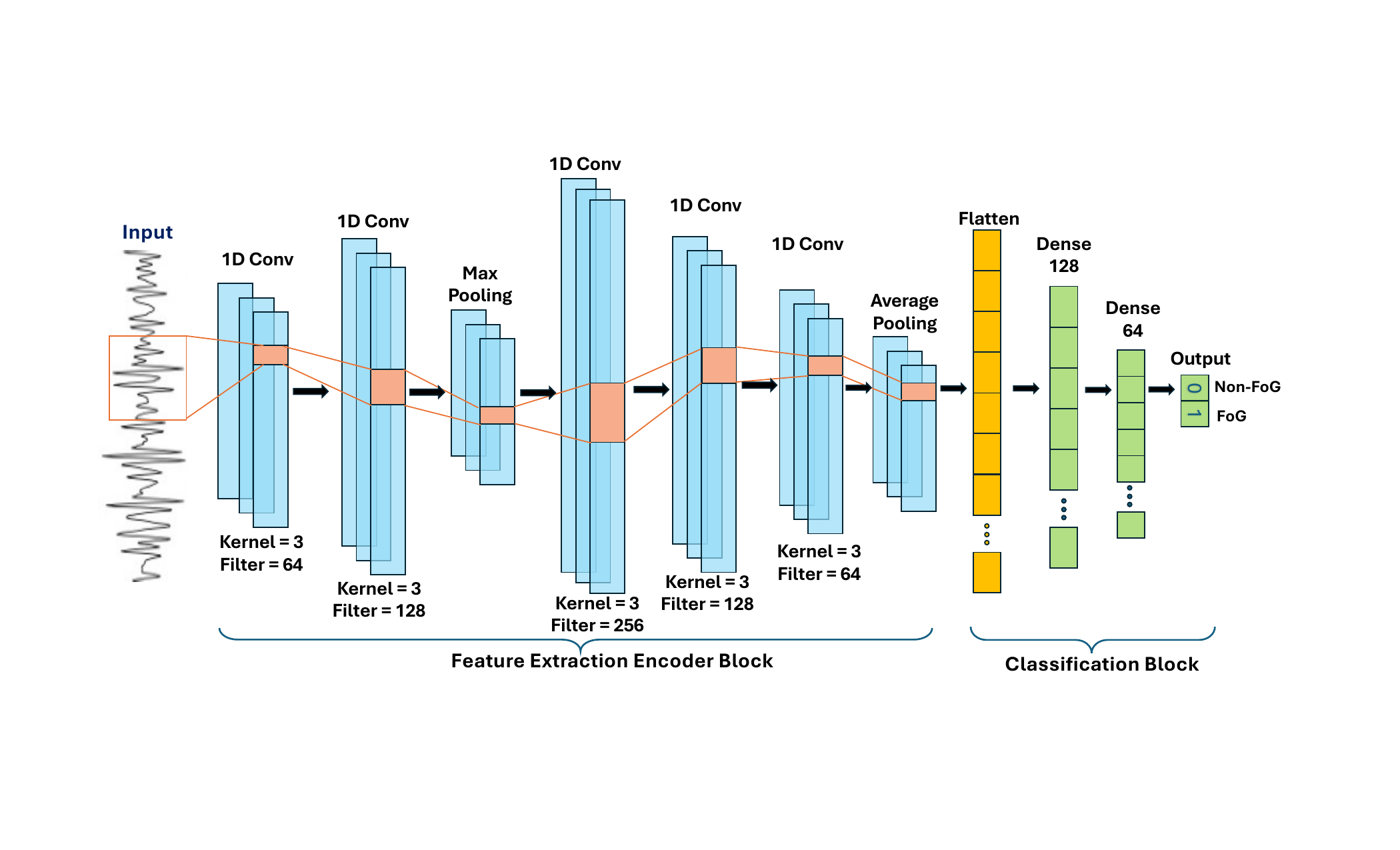}
    \vspace{-2mm}
    \caption{Architecture of stacked 1D CNN model. Input to the model is 3-minute raw sensor data.}
    \label{fig:model}
    \vspace{-2mm}
\end{figure}

For our downstream task, we devise a 5-layer 1D Convolutional Neural Network (CNN) architecture in LIFT-PD (Fig.~\ref{fig:model}), allowing us to use raw sensor data without extensive feature engineering. The model consists of an encoder block for feature extraction and a classification block for FoG detection. 
The encoder block has five 1D convolutional layers (filters: 64, 128, 256, 128, 64; kernel size: 3; ReLU activation) with max-pooling (pool size: 2) after the second layer and global average pooling (pool size: 2) after the fifth layer to downsample feature maps, reducing complexity. The flattened output is passed through an MLP with two dense layers (units: 128, 64; ReLU activation; dropout: 0.4 after the first layer) for binary classification, with a final dense layer (unit: 1; sigmoid activation) providing the FoG detection output.

The model was trained in two phases: pre-training and fine-tuning. In the pre-training phase, the encoder was trained for 70 epochs with a batch size of 64, using the Adam optimizer set to a learning rate of 0.01 and a decay of 0.001. 
During the fine-tuning phase, 
the additional dense layers were randomly initialized. The model was fine-tuned on labeled data for 40 epochs, maintaining the same batch size but with a reduced learning rate of 0.0001
for the classification task. The use of a lower learning rate (0.0001) during the fine-tuning phase is a common practice when working with pre-trained models. This lower learning rate helps preserve the learned features from the pre-training phase and ensures gradual adjustments, avoiding the drastic loss of previously learned representations.

All experiments (pre and post-processing) were performed on a computer with an Apple M2 Pro chip, which includes a 16-core Neural Engine, and 16 GB of unified memory. Training, optimization, and testing of the classification model were performed in Python (3.8), using the Keras (2.4), and TensorFlow (2.3) libraries.

\begin{algorithm}
    \caption{Activity Threshold Optimization (ATO) Algorithm for Prolonged Battery Life}
    \label{alg:activity-threshold}
    \begin{algorithmic}[1]
     \State {\bfseries Input:} $X$: Set of inference data frames, 
    $[\alpha_{start}, \alpha_{final}]$: Initial and final threshold, 
    $P_0$: Baseline performance metric,
    $f(\cdot)$: Inference Model,\
    $A(\cdot)$: returns magnitude for $x_i \in X$
    \State {\bfseries Output:}  $\alpha_{opt}:$ Optimal activity threshold, $X_{opt}$: Active frames for ($\alpha_{opt}$)
    \State {\bfseries Begin}
    \State initialize $\alpha \gets \alpha_{start}$

        \While{$\alpha \leq \alpha_{final}$}
        \State  $X_{active} = \{x_i \in X \mid A(x_i) \geq \alpha\}$  \Comment{Set of active frames}\;
        
        \State $N_{\alpha} \gets {} \mid X_{active}\mid$ \Comment{Size of active frames}\;
        \State $P \gets f(X_{\alpha})$

        \If{$|P - P_0| \le \theta$}
            \State $\alpha \gets \alpha + \delta\alpha$
        \Else
            \State $\alpha_{\text{opt}} \gets \alpha$
            \State $X_{opt} \gets \{x_i \in X \mid A(x_i) \geq \alpha_{opt}\}$ 
            \State \textbf{break}
        \EndIf
        
        \EndWhile
  
        \State \Return $\alpha_{opt}, X_{opt}$  \Comment{optimal threshold, activity frames}
    \end{algorithmic}
\end{algorithm}

\subsection{Performance Measures}
To evaluate the performance in \textcolor{red}{FoG} detection at window-level of our proposed framework and those reproduced from the state-of-the-art approaches, commonly used metrics were calculated and reported. 

\begin{figure}[!htb]
\centering
\includegraphics[width=0.55\linewidth, trim={10 40 10 35},clip]{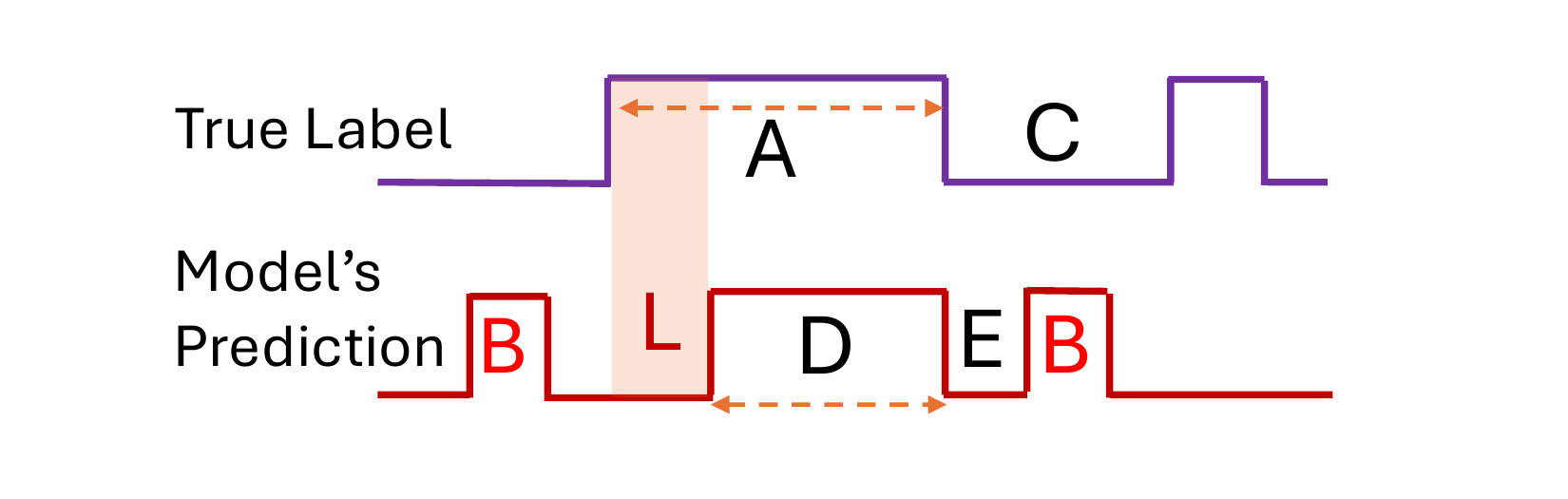}
\caption{A: FoG Episode, B: False Positive, C: Normal Episode, D: Detected FoG Episode (TP), E: Detected Normal Episode (TN), L: Latency}
\label{fig:groundtruth}
\vspace{-2mm}
\end{figure}

In this binary classification problem (FoG or non-FoG), performance metrics include sensitivity, specificity, F1 score, and AUC of receiver operating characteristics. We extensively evaluated our proposed system using these metrics, comparable to other state-of-the-art methods. True positives (TP) are correctly identified FoG windows, while false positives (FP) are non-FoG windows incorrectly labeled as FoG. False negatives (FN) are real \textcolor{red}{FoG} windows not recognized, and true negatives (TN) are correctly classified non-FoG instances. Fig.~\ref{fig:groundtruth} schematically describes these metrics.

Sensitivity/Recall measures the ratio of correctly detected \textcolor{red}{FoG} windows and specificity measures the ratio of correctly identified non-FoG windows. Precision evaluates the model's ability to avoid false positives. The F1 score is the harmonic mean of sensitivity and precision which is used to assess performance on the imbalanced dataset~\cite{10054906}.

\vspace{-3mm}
\subsection{FoG Detection Post-processing}
For further evaluation of our LIFT-PD framework, a post-processing analysis was performed using predictions and class labels. The performance of \textit{true FoG episode (TFE)} detection was performed by analyzing groups of consecutive windows with the presence of freezing episodes, besides the window-level FoG detection.

\begin{table*}[t]
  \centering
  \caption{Performance of proposed \textit{LIFT-PD} framework.  Key Metrics Defined: PREC: Precision, REC: Recall, SENS: Sensitivity, ACC: Accuracy, SPEC: Specificity, DFE: Detected FoG Episodes, DFW: Detected FoG Windows, STD: Standard Deviation. Performance metrics for the baseline supervised model are presented within parentheses ().}
  \label{tab:performance-metrics}
  \setlength{\tabcolsep}{3.5pt} 
\renewcommand{\arraystretch}{1} 
  \begin{tabular}{c||*{6}{c}|cc}
    \toprule
    Group & Prec & Rec/Sens/DFW & F1 Score & Acc. & Spec. & Loss & DFE & DFW  \\
    \midrule
    1 & 0.66 (0.61) & 0.82 (0.82) & 0.73 (0.70) & 0.81 (0.77) & 0.81 (0.75) & 0.19 (0.23) & 86.65\% (86.7\%) & 81.6\% (82.0\%)  \\
    2 & 0.82 (0.77) & 0.86 (0.86) & 0.84 (0.81) & 0.84 (0.82) & 0.83 (0.78) & 0.16 (0.19) & 89.35\% (90.1\%) & 86.1\% (86.1\%)  \\
     \midrule
    Avg. & 0.74 (0.69) & 0.84 (0.84) & 0.79 (0.76) & 0.825 (0.79) & 0.82 (0.77) & 0.18 (0.21) & 88.00\% (88.4\%) & 83.85\% (84.1\%)  \\
    \midrule
    Min & 0.66 (0.61) & 0.82 (0.82) & 0.73 (0.70) & 0.81 (0.77) & 0.81 (0.75) & 0.16 (0.19) & 86.65\% (86.7\%) & 81.60\% (82.0\%)  \\
    Max & 0.82 (0.77) & 0.86 (0.86) & 0.84 (0.81) & 0.84 (0.82) & 0.83 (0.78) & 0.19 (0.23) & 89.35\% (90.1\%) & 86.10\% (86.1\%)  \\
    STD & 0.11 (0.11) & 0.03 (0.03) & 0.08 (0.08) & 0.02 (0.03) & 0.01 (0.02) & 0.02 (0.03) & 1.91 (2.4) & 3.18 (2.9) \\
    \bottomrule
  \end{tabular}
  \vspace{-3mm}
\end{table*}

\subsubsection{FoG Episode}
In our study, we assessed the detection of FoG episodes—defined as consecutive windows labeled as FoG—by computing both the percentage of episodes detected and the proportion of FoG accurately identified within each episode. An episode is considered detected if at least one window within it is correctly identified. FoG episodes were categorized into three groups: short ($<6$ seconds), medium ($6-12$ seconds), and long ($>12$ seconds).

Then we calculated the percentage of FoG episodes detected across the entire dataset and for each duration group. We also measured the proportion of accurately detected FoG windows within each episode (Fig.~\ref{fig:groundtruth}D), defined as $D_{FoG}(\%) = \frac{n_{\text{detected}}}{n_{\text{total}}}$, where \(n_{detected}\) represents the number of detected FoG windows, and \(n_{total}\) is the total FoG windows in that episode.


For false FoG episodes, we calculated the minimum distance between each falsely detected episode (Fig.~\ref{fig:groundtruth}B)  and the nearest true FoG window (Fig.~\ref{fig:groundtruth}A) to understand the proximity of false positives to actual FoG occurrences. 
Finally, we evaluated FoG detection latency, defined as the time difference between the onset of an actual FoG episode and the detected FoG episode (Fig.~\ref{fig:groundtruth}L). This metric reflects the algorithm’s responsiveness in identifying FoG events, which is crucial for the timely intervention and management of PD patients.

\subsubsection{Computational Complexity}


To evaluate the feasibility of deploying our LIFT-PD framework on low-resource wearable devices, we performed several analyses. We assessed training and testing times for different input sizes to identify the optimal training size for the self-supervised pre-text task using varying amounts of unlabeled data. Post-training, we measured inference times across different input sizes to gauge real-time performance.

We also analyzed memory requirements for storing input sensor data and model parameters, ensuring suitability for resource-constrained devices. To optimize power consumption and enable long-term in-home monitoring, we introduced an Activity Threshold Optimization (ATO) algorithm (Algorithm ~\ref{alg:activity-threshold}),
activating the FoG detection model only during active periods. 
Assuming that the performance function $P$ and the computation of active windows $N_{\alpha}$ can be performed in constant time, the overall runtime excluding the inference model is $O(N \cdot \frac{\alpha_{max}}{\delta\alpha})$. Including the inference model's runtime $O(M)$, where $M$ is the process time of a single active window, the adjusted complexity becomes $O((M+N) \cdot \frac{\alpha_{max}}{\delta\alpha})$.
In the best-case scenario, where the optimal threshold $\alpha_{opt}$ is found in the first iteration, the runtime is $O(N)$. 







%

\begin{table*}[h]
    \centering
    \caption{Performance comparison across diverse groups. Performance metrics for the baseline supervised model are presented within parentheses ().}
    \label{tab:diversity-performance}
    \renewcommand{\arraystretch}{1.2} 
    
    \begin{tabular}{c|*{5}{c|}c}
    \toprule
         Group & Test Group & Pre & Rec & F1 & Spec & Acc\\
         \midrule
         
        \multirow{2}{*}{\makecell{Sevirity\\(40)}} &Severe &\makecell{0.71 (0.7)} &\makecell{0.81 (0.79)} &\makecell{0.75 (0.74)} &\makecell{0.79 (0.79)} &\makecell{0.8 (0.79)}\\

        &Mild (14) & 0.78 (0.78) & 0.81 (0.81) & 0.80 (0.79) & 0.85 (0.84) & 0.83 (0.83)\\
         \hline

         \multirow{2}{*}{\makecell{Gender\\(40)}} &\makecell{Female} &\makecell{0.78 (0.67)} & \makecell{0.75 (0.77)} & \makecell{0.76 (0.72)} & \makecell{0.84 (0.71)} & \makecell{0.80 (0.77)} \\
         
         &\makecell{Male (32)} & \makecell{0.71 (0.74)} & \makecell{0.67 (0.64)} & \makecell{0.69 (0.68)} & \makecell{0.83 (0.86)} & \makecell{0.77 (0.77)} \\
        \hline
         
         \multirow{2}{*}{\makecell{Age\\(40)}} &\makecell{Old (20)} & \makecell{0.80 (0.75)} & \makecell{0.71 (0.71)} & \makecell{0.75 (0.74)} & \makecell{0.86 (0.82)} & \makecell{0.80 (0.78)} \\
         
         &\makecell{Mid-age} & \makecell{0.70 (0.66)} & \makecell{0.80 (0.79)} & \makecell{0.74 (0.72)} & \makecell{0.82 (0.79)} & \makecell{0.81 (0.79)} \\
        \hline

        \multirow{2}{*}{\makecell{Medication}} &\makecell{On} & \makecell{ 0.82 (0.78)} & \makecell{0.79 (0.72)} & \makecell{0.81 (0.75)} & \makecell{0.84 (0.82)} & \makecell{0.86 (0.81)} \\
         
         &\makecell{Off} & \makecell{0.78 (0.70)} & \makecell{0.8 (0.74)} & \makecell{0.79 (0.72)} & \makecell{0.79 (0.72)} & \makecell{0.82 (0.78)} \\
        \hline

         \multirow{2}{*}{\makecell{Random\\(40)}} &\makecell{1 (20)} & \makecell{0.66 (0.61)} & \makecell{0.82 (0.82)} & \makecell{0.73 (0.70) } & \makecell{0.81 (0.75)} & \makecell{0.81 (0.77)} \\
         
         &\makecell{2 (20)} & \makecell{0.82 (0.77)} & \makecell{0.86 (0.86)} & \makecell{0.84 (0.81)} & \makecell{0.83 (0.78)} & \makecell{0.84 (0.82)} \\
         
         \bottomrule
    \end{tabular}
    \vspace{-2mm}
\end{table*}

\section{Results}

\subsection{Performance Analysis}
\label{sub-sec:result.performance}

We evaluated the performance of the LIFT-PD framework using mainly sensitivity, and specificity along with some other matrices (percentage of detected episode, latency, precision, F1 score, and AUC of the ROC curve), comparing it against a baseline supervised model with the same architecture and parameters. This comprehensive evaluation highlights the benefits and potential limitations of our self-supervised learning (SSL) approach for real-time FoG detection.

Table~\ref{tab:performance-metrics} summarizes the metrics, showing that the SSL model achieved notable improvements: a 7.25\% increase in average precision, 4.4\% in accuracy, and 6.5\% in specificity compared to the baseline supervised model. Importantly, the SSL model maintained consistent recall/sensitivity (84\%) with the baseline, ensuring that the detection of FoG episodes was not compromised despite reduced supervision. The F1 score, which balances precision and recall, was higher by about 3.95\% in the SSL model, indicating better overall performance in detecting FoG episodes.

Group 2 outperformed Group 1 in all metrics. Group 1's lower precision (0.66) was due to more false positives from a higher proportion of FoG events. The DHW technique with SSL mitigated data imbalance, enhancing performance despite class imbalances in Group 1.
\begin{figure}[htb]
\vspace{-4mm}
\centering
\includegraphics[width=0.65\linewidth, trim={0 10 30 20},clip]{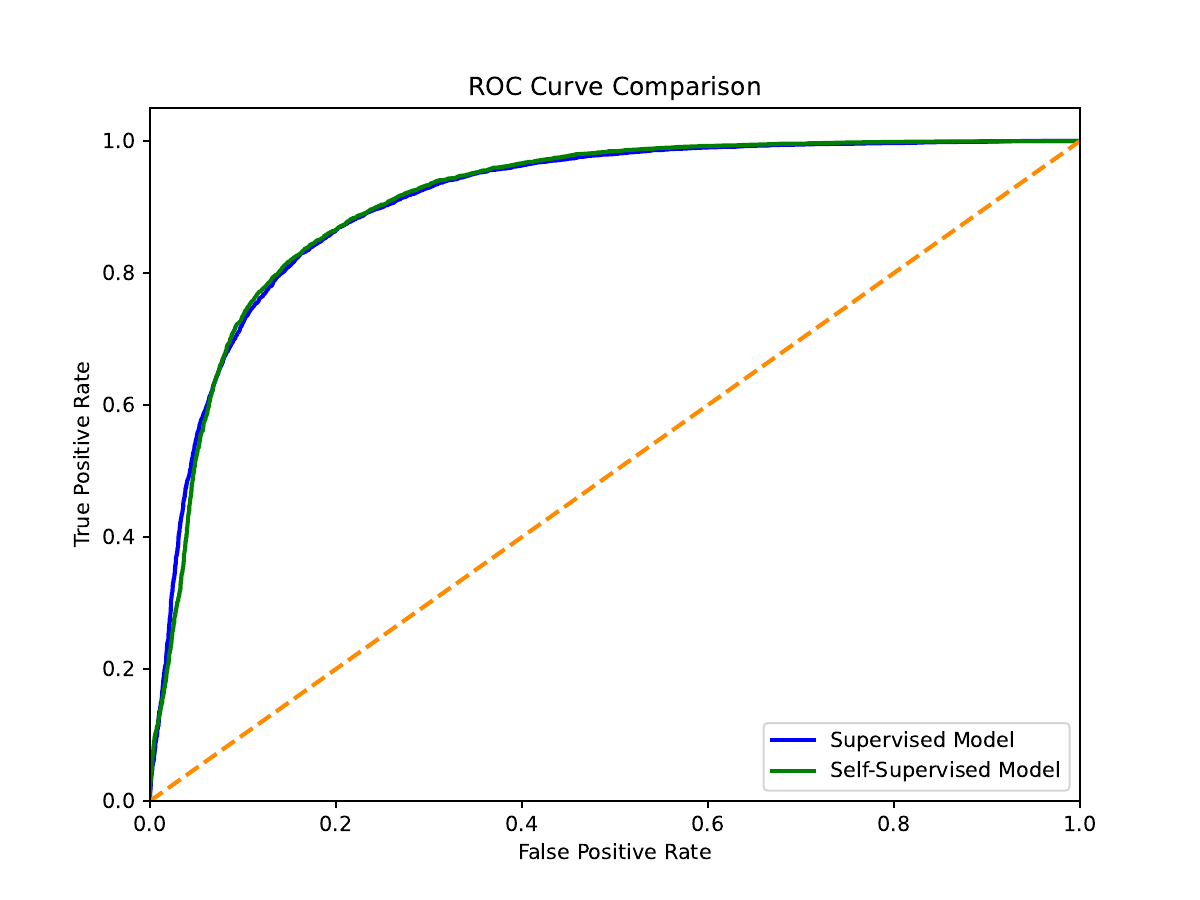}
\vspace{-4mm}
\caption{Receiver Operating Characteristic (ROC) curves for supervised and self-supervised models}
\vspace{-2mm}
\label{fig:roc}
\end{figure}
The ROC curves in Fig.~\ref{fig:roc} showed that the SSL model had a slightly larger area under the curve (0.908) than the supervised model (0.9078), indicating better FoG classification performance.
The close proximity of the AUC values between the two models highlights the robustness of our SSL approach in achieving comparable performance to the supervised baseline, despite leveraging limited labeled data during training.

\begin{table}[h]
\color{black}
\small
\centering
\caption{\textcolor{black}{Comparison between the proposed Self-Supervised Learning (SSL) model and two Semi-Supervised Learning baselines—Mean-Teacher (MT) and $\Pi$ model—using the same 1-D CNN backbone. Metrics are averaged across groups; episode-level results (DFE) show detection rates for different FoG episode durations.}}
\label{tab:semi-SL_performance-metrics}
\setlength{\tabcolsep}{4.5pt}
\renewcommand{\arraystretch}{1.15}

\begin{tabular}{c||ccc|ccc|ccc|ccc|ccc}
\toprule
\multirow{2}{*}{\textbf{Group}} 
& \multicolumn{3}{c|}{\textbf{Precision}} 
& \multicolumn{3}{c|}{\textbf{Rec/Sens/DFW}} 
& \multicolumn{3}{c|}{\textbf{F1 Score}} 
& \multicolumn{3}{c|}{\textbf{Accuracy}} 
& \multicolumn{3}{c}{\textbf{Specificity}}\\
& \cellcolor{sslcolor}SSL & \cellcolor{baseline2}MT & \cellcolor{baseline1}$\Pi$
& \cellcolor{sslcolor}SSL & \cellcolor{baseline2}MT & \cellcolor{baseline1}$\Pi$
& \cellcolor{sslcolor}SSL & \cellcolor{baseline2}MT & \cellcolor{baseline1}$\Pi$
& \cellcolor{sslcolor}SSL & \cellcolor{baseline2}MT & \cellcolor{baseline1}$\Pi$
& \cellcolor{sslcolor}SSL & \cellcolor{baseline2}MT & \cellcolor{baseline1}$\Pi$\\
\midrule
1 & \cellcolor{sslcolor!50}\textbf{0.66} & 0.65 & 0.66 
  & \cellcolor{sslcolor!50}\textbf{0.82} & 0.75 & 0.73 
  & \cellcolor{sslcolor!50}\textbf{0.73} & 0.70 & 0.70 
  & \cellcolor{sslcolor!50}\textbf{0.81} & 0.79 & 0.80 
  & \cellcolor{sslcolor!50}0.81 & 0.81 & \textbf{0.84} \\
2 & \cellcolor{sslcolor!50}\textbf{0.82} & 0.76 & 0.80 
  &\cellcolor{sslcolor!50} \textbf{0.86} & 0.74 & 0.81 
  & \cellcolor{sslcolor!50}\textbf{0.84} & 0.75 & 0.80 
  & \cellcolor{sslcolor!50}\textbf{0.84} & 0.77 & 0.82 
  & \cellcolor{sslcolor!50}\textbf{0.83} & 0.80 & 0.82 \\
\midrule
\textbf{Average} & \cellcolor{sslcolor!60}\textbf{0.74 }& 0.71 & 0.74 
  & \cellcolor{sslcolor!50}\textbf{0.84} & 0.75 & 0.77 
  & \cellcolor{sslcolor!50}\textbf{0.79} & 0.72 & 0.75 
  & \cellcolor{sslcolor!50}\textbf{0.83} & 0.78 & 0.81 
  & \cellcolor{sslcolor!50}0.82 & 0.81 & \textbf{0.83} \\
\bottomrule
\end{tabular}

\vspace{2mm}

\begin{tabular}{ccc|ccc|ccc}
\toprule
\multicolumn{9}{c}{\textbf{\% of Detected FoG Episodes (DFE)}}\\
\midrule
\multicolumn{3}{c|}{\makecell{\textbf{Small (0--6 s)}}} &
\multicolumn{3}{c|}{\makecell{\textbf{Medium (6--12 s)}}} &
\multicolumn{3}{c}{\makecell{\textbf{Large ($>$12 s)}}}\\
\cellcolor{sslcolor}SSL & \cellcolor{baseline2}MT & \cellcolor{baseline1}$\Pi$ &
\cellcolor{sslcolor}SSL & \cellcolor{baseline2}MT & \cellcolor{baseline1}$\Pi$ &
\cellcolor{sslcolor}SSL& \cellcolor{baseline2}MT & \cellcolor{baseline1}$\Pi$ \\
\midrule
\cellcolor{sslcolor!50}\textbf{83.3}\% & 77.7\% & 79.7\% &
\cellcolor{sslcolor!50}\textbf{100}\% & 96.0\% & 97.0\% &
\cellcolor{sslcolor!50}\textbf{100}\% & 99.2\% & 99.2\% \\
\bottomrule
\end{tabular}

\end{table}

\subsubsection{\textcolor{black}{Compare with Baselines}}
\textcolor{black}{
We also compared the performance of our self-supervised learning (SSL) approach with two semi-supervised learning (Semi-SL) baselines for FoG detection. To isolate the benefit of our representation-learning strategy, we replaced the SSL pre-training stage in LIFT-PD with (i) a $\Pi$‑model–based semi-supervised pipeline~\cite{FixMatch} and (ii) a Mean-Teacher (MT) variant~\cite{tarvainen2017mean}, while keeping the backbone network, DHWT segmentation, and opportunistic inference module identical across models. Both Semi-SL methods leverage unlabeled data through consistency regularization, yet differ in how pseudo-labels are formed: the $\Pi$-model relies on confidence-filtered student predictions, whereas the MT uses an exponential-moving-average (EMA) teacher to provide soft, stable targets.
}

\textcolor{black}{
Although both baselines improved over purely supervised training, they required a larger fraction of labeled data ($\approx$65–75\% of the training set, vs. 40\% for LIFT‑PD) for stable pseudo-label convergence. Due to the sparsity and variability of FoG episodes, early pseudo-labels in the $\Pi$-model tended to be noisy, while the MT showed better stability but lower sensitivity under limited label ratios. In contrast, our SSL pre-text task learns label-agnostic temporal structure by reconstructing masked sensor segments, enabling the encoder to generalize from all recordings without relying on tentative labels. As a result, LIFT-PD achieved higher recall (0.84 vs. 0.75 [MT] / 0.77 [$\Pi$]) and F1-score (0.79 vs. 0.72 [MT] / 0.75 [$\Pi$]), while matching or exceeding both baselines in precision, accuracy, and specificity (Table~\ref{tab:semi-SL_performance-metrics}).
}

\subsubsection{Generalizibility of DHWT}
\label{subsub-sec:dhwt-results}
\begin{table}[h]
    \centering
    \caption{Performance comparison of DHWT across datasets.  The second row shows relative improvement over the baseline.}
    \label{tab:dhwt-performance}
    \setlength{\tabcolsep}{2.2pt} 
    \renewcommand{\arraystretch}{1.5} 
    \begin{tabular}{c|*{4}{c} cc}
    \toprule
         Dataset & Train Time   & Test Time & Pre & F1 & Spec & Acc\\
         \midrule
         
         tdcs &\makecell{1.8s\\$\uparrow$88\%} & \makecell{0.94s\\(0.92s)} &\makecell{0.74\\$\uparrow$7.2\%} &\makecell{0.8\\$\uparrow$5.3\%} &\makecell{0.82\\$\uparrow$6.5\%} &\makecell{0.83\\ $\uparrow$5.1\%}\\
         \hline
         
         Daphnet     &\makecell{0.14s \\$\uparrow$42\%} &\makecell{0.134s\\(0.132s)}       &\makecell{0.55 \\ $\uparrow$161\%}   &\makecell{0.44\\$\uparrow$42\%}   &\makecell{0.94\\$\uparrow$108\%} &\makecell{0.87\\$\uparrow$71\%}\\
         \hline
         
         MotionSense &\makecell{0.08 \\ $\uparrow$142\%} &\makecell{0.022s \\ (0.024s)} &\makecell{1 \\ $\uparrow$3.1\%} &\makecell{0.99 \\ $\uparrow$2.5\%} &\makecell{1 \\ $\uparrow$1.2\%} &\makecell{0.99 \\ $\uparrow$1.22\%}\\
         \bottomrule
    \end{tabular}
    \vspace{-2.8mm}
\end{table}
We expanded our analysis to Daphnet~\cite{daphnet_freezing_of_gait_245}  and two activities from MotionSense~\cite{Malekzadeh:2019:MSD:3302505.3310068} (as a binary classification), in addition to the primary tdcs dataset. The results show an increase in training time across the datasets due to class imbalance handling, while the inference time remained stable. DHWT significantly improved precision, F1-score, and specificity in datasets with class imbalance, confirming its effectiveness across diverse conditions. Daphnet showed a 161\% boost in precision and a 108\% improvement in specificity. These results support the generalizability of DHWT without compromising computational efficiency (Table~\ref{tab:dhwt-performance}).

\subsubsection{Robustness Across Diverse Subjects}
\label{subsub-sec:diveristy-results}

To ensure generalizability, we split the dataset(N=40) into diverse groups based on key characteristics such as severity (Mild N=14), gender (Male N=32) and age (Mid-age N=20). These groups were deliberately formed to include outliers and reflect real-world diversity, ensuring that the model can handle a variety of subject profiles without overfitting to any particular population. 
We observed performance improvements in F1 scores, accuracy and specificity in all groups, particularly in mild cases and older age groups, confirming that the method works effectively even with varied patient characteristics. 

Regarding the medication status, we evaluated the model separately on data from patients in their ``On" and ``Off" medication states. Patients in the ``On" state generally show fewer motor symptoms due to the effects of dopaminergic treatment, while the ``Off" state is marked by more pronounced motor symptoms, including fluctuations.  Interestingly, when both ``On" and ``Off" states were combined during training, the model demonstrated superior results by better capturing the intra-patient variability between medicated and unmedicated states (Random group-2). This suggests that LIFT-PD can adapt to fluctuating motor symptoms effectively, providing accurate detection across different medication states, which is critical for real-world clinical monitoring and intervention planning.
\subsection{Clinical Correlation of Predicted FoG Scores}
To assess the clinical relevance of LIFT-PD's predictions, we computed correlation coefficients between the model’s FoG scores and standard clinical metrics. Here, FoG scores were defined as the per-subject proportion of detected FoG windows during test sessions. This metric provides a summary-level indicator of how frequently the model detected FoG-related patterns in each participant.
We found that the FoG scores were significantly correlated with NFOGQ scores (Pearson’s r = 0.72, p $<$ 0.01) and UPDRS-III Off scores (r = 0.63, p $<$ 0.05), both widely used clinical markers of disease severity and motor impairment. These results indicate strong alignment between the model's predictions and clinical assessments, supporting the model's translational potential for patient monitoring.

\subsection{Statistical Analysis of Performance Differences}
\begin{table}[h]
\vspace{-2mm}
    \centering
    \caption{The paired t-test results comparing LIFT-PD to the supervised models}
    \label{tab:paired_ttest}
    \setlength{\tabcolsep}{2.1pt} 
    \renewcommand{\arraystretch}{1.1} 
    \begin{tabular}{c|ccp{5cm}}
    \toprule
         \textbf{Metric}     &\textbf{t-statistic }   &\textbf{p-value }   &\textbf{Interpretation}\\
         \midrule
         Precision      &3.18    &0.011   &Statistically significant improvement with LIFT-PD (p$<$0.05)\\
         Recall         &1.87    &0.094   &Not statistically significant (p$ > $0.05), though LIFT-PD shows a positive trend\\
         F1 Score       &4.20    &0.0023   &Highly significant improvement in F1 score with LIFT-PD.\\
         Specificity    &2.74    &0.023   &Significant improvement in reducing false positives (p$ < $0.05).\\
         Accuracy       &4.27    &0.0021   &Highly significant improvement in overall accuracy\\
         \bottomrule
    \end{tabular}
    \vspace{-2.8mm}
\end{table}
To statistically validate the improvements observed with LIFT-PD over the supervised baseline, we conducted paired t-tests across multiple performance metrics. The results show statistically significant improvements in precision (p = 0.011), F1-score (p = 0.0023), specificity (p = 0.023), and accuracy (p = 0.0021), confirming that LIFT-PD outperforms the supervised model in key areas relevant to clinical application. Although the improvement in recall (p = 0.094) was not statistically significant at the 5\% level, it still showed a positive trend, suggesting consistent performance across diverse patient groups. These findings support the robustness and reliability of the proposed approach.

\subsection{Compare with State-of-the-Art Models}
Table~\ref{tab:performance-compare} compares the performance of our LIFT-PD model with state-of-the-art methods for detecting FoG in PD patients. For fairness, all models were implemented from scratch and evaluated on the same dataset using consistent experimental protocols. Although the Multi-head CNN~\cite{10.1016/j.artmed.2022.102459} achieves the highest detection rates for FoG episodes (97.27\%) and windows (94.64\%), its precision (0.545) and specificity (0.491) are low, indicating a high false positive rate. High false positives can reduce the effectiveness of cueing due to patients' adaptiveness to ``always on" interventions~\cite{ginis2018cueing}~\cite{cubo2004short}. The one-class classifier~\cite{naghavi2021towards} shows high precision (0.856) and specificity (0.891) but lower recall (0.716), missing many true FoG events, which can lead to inadequate monitoring and delayed interventions. The semi-supervised model~\cite{mikos2017real}  shows a recall around 12\% lower and specificity approximately 13\% lower than LIFT-PD, making it less suitable for in-home monitoring with a single accelerometer.

\begin{table}[h]
\vspace{-2mm}
    \centering
    \caption{Comparison with state-of-the-art models}
    \label{tab:performance-compare}
    \renewcommand{\arraystretch}{1.3} 
    \begin{tabular}{c|*{3}{c} cc}
    \toprule
         Study & DFE   & Rec/Sens/DFW & Prec & F1 Score & Spec.\\
         \midrule
         One Class Classifier~\cite{naghavi2021towards}         &90.8\% &71.6\%       &0.86   &0.77   &0.89\\
         Semi-Supervised Model~\cite{mikos2017real} & -- & 72.3\% &--&-- & 0.71 \\
         Multi-head CNN~\cite{10.1016/j.artmed.2022.102459}& 97.3\%&94.6\%      &0.56   &0.68   &0.49\\
         \hline\hline
         \textbf{LIFT-PD}          &88\%   &\underline{84\%}         &\underline{0.74}   &\textbf{0.79}   &\underline{0.82} \\
         \bottomrule
    \end{tabular}
    \vspace{-2.8mm}
\end{table}
Our LIFT-PD model achieves a balanced performance with 88\% episode detection, 83.85\% window detection, 0.74 precision, 0.84 recall, 0.79 F1-score, and 0.82 specificity. These results show that LIFT-PD achieves comparable performance to the supervised methods while being more suitable for real-time wearable deployment with limited computational resources for remote monitoring.

\subsection{Duration Based FoG Episode}
\label{sub-sec:result.epsisode}
Table~\ref{tab:fog-episodes} presents the detection rates for FoG episodes and windows, along with associated latency metrics, grouped into short, medium, and long durations. Our SSL model's performance is compared with a baseline supervised model (metrics in parentheses).

The SSL model shows robust performance across all durations, with detection rates increasing as the episode lengthens. This trend is mirrored in the baseline model’s performance, indicating a consistent improvement across different modeling approaches. For short episodes, the detection rate is 83.3\% for episodes and 68\% for windows, with an average latency of 2.42±0.45 seconds. For long episodes, these rates rise to 100\% for episodes and 91.1\% for windows, with a latency of 2.64\(\pm\)1.45 seconds. The increasing detection rates for longer episodes are due to the more prominent and persistent FoG characteristics, such as tremors and shuffling gait, which the SSL model effectively captures. The increasing detection rates for longer episodes are due to the more prominent and persistent FoG characteristics, such as tremors and shuffling gait, which LIFT-PD effectively captures.
\begin{table}[!htb]
\centering
\caption{Duration-based FoG episode analysis.}
\label{tab:fog-episodes}

\setlength{\tabcolsep}{4pt} 
\renewcommand{\arraystretch}{1} 
\begin{tabular}{@{}l|ccc@{}}
\toprule
 \multirow{2}{*}{\makecell{Duration}} & \multicolumn{3}{c}{\makecell{FoG Episodes \& windows detection rate (\%) with \\ latency (s) and standard deviation} }  \\ \cmidrule(l){2-4} 
         & \makecell{\textbf{Small, 0--6 s }}& \makecell{\textbf{Medium, 6--12 s}} & \makecell{\textbf{Large, \textgreater{}12 s}} \\ \midrule
\makecell{FoG Episodes}  & \makecell{83.3\% \\ (82.8\%)}    & \makecell{100\% \\ (98.8\%)}     & \makecell{100\% \\ (100\%)}   \\ \hline 
\makecell{FoG Windows}    & \makecell{68\% \\ (71.4\%)}     & \makecell{81.9\% \\ (84.1\%)}   & \makecell{91.1\% \\ (92.6\%)}  \\ \hline  
\makecell{Avg. Latency \(\pm\) SD}  & \makecell{2.42\(\pm\)0.45 s  \\ (2.38\(\pm\)0.45 s)}     & \makecell{2.6\(\pm\)0.96 s \\ (2.5\(\pm\)0.82 s)}       & \makecell{2.64\(\pm\)1.45 s \\ (2.6\(\pm\)1.29 s)}  \\ \hline
\makecell{Max. Latency} & \makecell{ 4.5 s \\ (4.5 s)} & \makecell{6 s \\ (5.25 s)} &  \makecell{9.75 s \\ (9.75 s)} \\
\bottomrule
\end{tabular}
\vspace{-2mm}
\end{table}

Latency, representing the detection delay from episode onset, slightly increases with duration. Small episodes have an average latency of 2.42±0.45 seconds, while long episodes reach 2.64±1.45 seconds. Maximum latency values increase from 4.5 seconds for small episodes to 9.75 seconds for long ones, reflecting the need to analyze more data over time to confidently detect FoG onset. Despite this latency increase, the detection accuracy for longer episodes remains superior, showing the SSL model's effectiveness in real-world applications by balancing latency and accuracy for reliable FoG detection across varying episode durations.

\textcolor{black}{
\textbf{\textit{Failure analysis for short FoG episodes:}} Short episodes (< 6 s) show a lower episode-level detection rate (83.3\%, Table~\ref{tab:fog-episodes}) mainly because their onsets and offsets often coincide with window boundaries. This reduces the proportion of FoG-dominant samples within a 3-s window, leading to ambiguous temporal cues and occasional delays. In addition, when the MAM threshold remains high at the start of a movement bout, model activation may be deferred, slightly delaying the first positive prediction. Future work will explore adaptive windowing (e.g., 2.5-s windows) and boundary-aware smoothing to improve responsiveness for brief episodes.
}

\subsection{Threshold Effect in Model Activation Module (MAM)}
\label{sub-sec:result.threshold}
\begin{figure}[!htb]
\vspace{-2mm}
\centering
\includegraphics[width=0.7\linewidth]{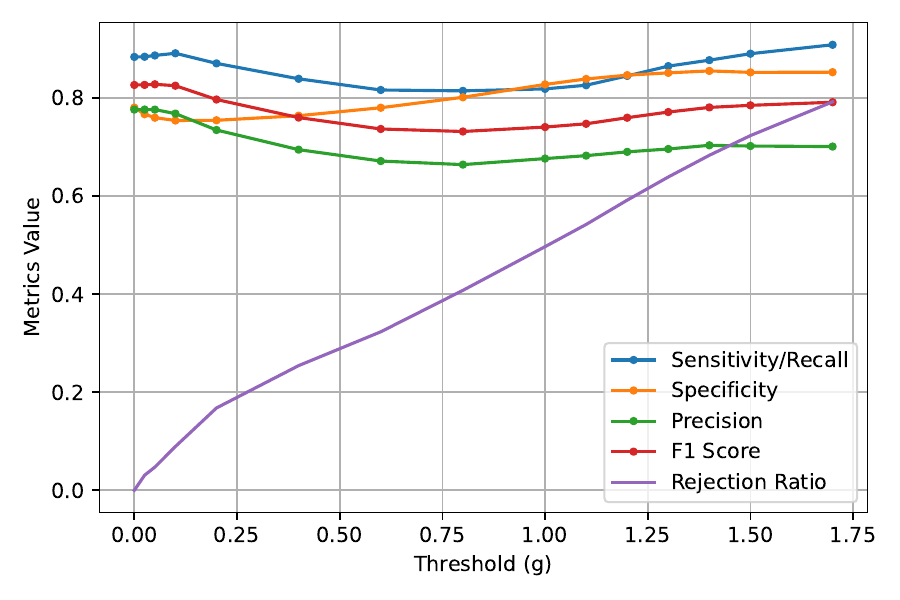}
\vspace{-5mm}
\caption{Impact of activity threshold on SSL performance}
\label{fig:sslmetric}
\end{figure}

To determine the optimal activity threshold for the MAM, we evaluated the impact of different threshold values on various performance metrics, as presented in Table~\ref{tab:mam-performance} and Fig.~\ref{fig:sslmetric}. 
As observed in Table~\ref{tab:mam-performance}, lowering the activity threshold (e.g., 0.0) results in higher sensitivity (0.884) and higher detected FoG episode (DFE) rate (0.92), indicating that more FoG events are detected. However, this comes at the cost of lower specificity (0.78) and a higher inference time (3.31 ms) due to the SSL model being activated more frequently, even during inactive periods. Conversely, increasing the activity threshold (e.g., 1.2) leads to higher specificity (0.846) and a lower rejection ratio (0.59), implying that fewer false positive detections occur during inactive periods. However, this improvement is accompanied by a slight decrease in sensitivity (0.845), DFE rate (0.809), and an increase in the number of missed FoG events.
\begin{table}[htbp]
\vspace{-2mm}
\centering
\caption{Effect of different activation thresholds on MAM}
\label{tab:mam-performance}
\setlength{\tabcolsep}{5pt} 
\begin{tabular}{cccccc}
\toprule
\makecell{Threshold}& Sens & Spec & DFE & \makecell{Rejection Ratio} & \makecell{Inference Time} \\
\midrule
0  & 0.884    & 0.78     & 0.92         & 0     & 3.31 ms \\
0.2     & 0.87     & 0.75     & 0.92    & 0.17  & 1.96 ms    \\ 
0.4     & 0.839     & 0.764     & 0.896 & 0.25  & 1.76 ms    \\
1.2		& 0.845	   & 0.846	    &0.809  & 0.59  & 1.1 ms \\																				
\bottomrule
\end{tabular}
\end{table}

\begin{figure}[!h]
    \centering
    \includegraphics[width=0.7\linewidth]{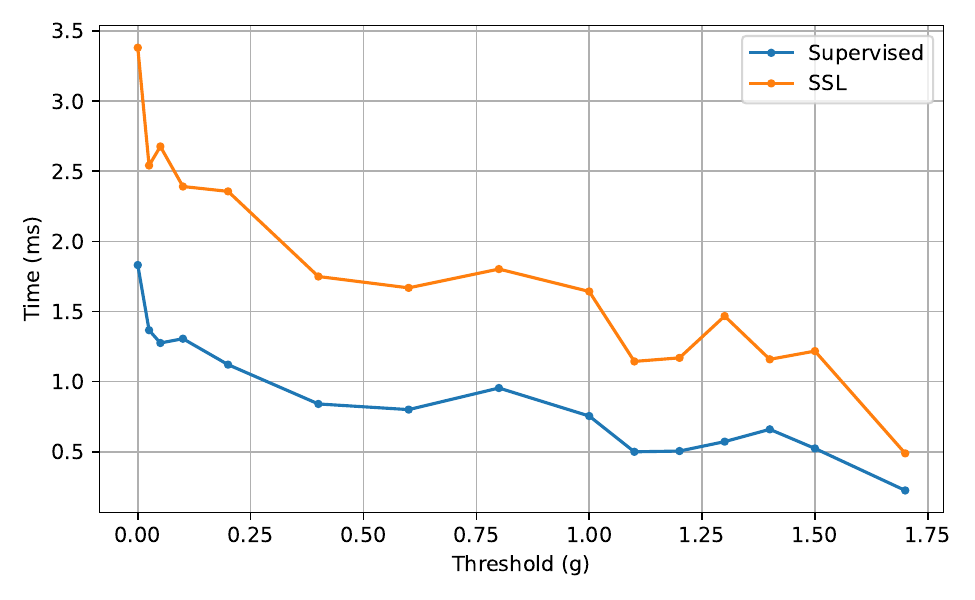}
    
    
    \vspace{-5mm}
    \caption{Effect of activity threshold on inference time}
    \label{fig:sslinference}
    \vspace{-2mm}
\end{figure}
Fig.~\ref{fig:sslmetric} shows how metrics like specificity and rejection ratio increase steadily with the threshold, while sensitivity and overall detection performance (F1 score) gradually decline. Careful selection of the threshold is crucial to balance the requirements of accurately detecting active periods while effectively rejecting inactive intervals to optimize computational resources.
Fig.~\ref{fig:sslinference} shows that as the activity threshold increases, the model execution time decreases for both supervised and self-supervised models. At a threshold of 1.2g, the SSL model's execution time reaches approximately 1.1 ms (Table~\ref{tab:mam-performance}, Fig.~\ref{fig:sslinference}).
This reduction is due to the MAM effectively filtering out inactive periods, reducing the need for the computationally intensive SSL model. The SSL model has slightly higher inference times than the supervised model due to an additional MLP layer on top of the pre-trained encoder. Despite the modest increase, the SSL model demonstrates enhanced robustness and superior performance using only 40-60\% of labeled training data. The average execution time per window is 0.0295 ms for the supervised model and 0.0379 ms for the SSL model.

\subsubsection{Battery-Life Estimation}
\textcolor{black}{
Battery-life estimation was derived from measured inference times obtained on the local hardware used for experimentation. 
Each activity-threshold setting (\( \alpha \)) yielded a distinct average inference time (Table~\ref{tab:mam-performance}), which was used as a proxy for energy consumption. 
Assuming energy consumption scales linearly with inference duration, the average compute energy can be modeled as
\[
E_{\text{avg}} = (1 - \rho) E_{\text{inf}} + E_{\text{base}},
\]
where \( \rho \) denotes the rejection ratio and \( E_{\text{inf}} \) corresponds to the measured inference energy (or time). 
At \( \alpha = 1.2\,g \), the rejection ratio reached \( 0.59 \), reducing the inference time from \( 3.31\,\text{ms} \) to \( 1.1\,\text{ms} \)---approximately a \( 2.4\times \) decrease in computational duty. 
This improvement translates to an estimated runtime extension from about 20 hours to over 48 hours on a comparable low-power wearable device. 
Because MAM operates on the mean-removed acceleration magnitude, it remains robust to sensor orientation, minor displacement, and brief motion artifacts.
}
\subsection{Effect of Window Size in Episode Detection}
\label{subsec:windowsize}
\begin{figure}[h]
\vspace{-2mm}
\centering
\includegraphics[width=0.65\linewidth]{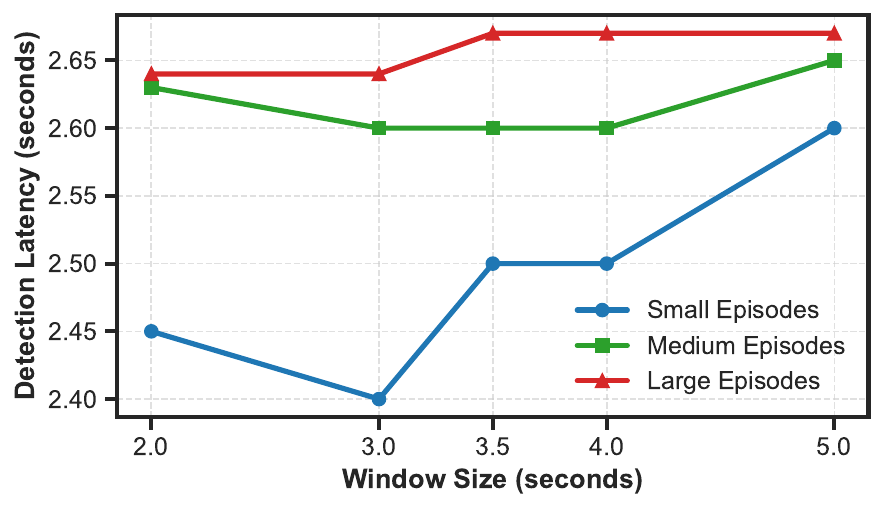}
\vspace{-2mm}
\caption{Impact of window size on FoG detection latency across different episode durations (Small, Medium, Large).}
\label{fig:window-size}
\vspace{-2mm}
\end{figure}

We conducted extensive validation by randomly taking a 10\% subset of the training data as a validation set. Our results indicated that longer windows, such as 4s or 5s, showed a delay in detecting FoG episodes, possibly due to including unnecessary gait segments not closely linked to FoG onset, in line with previous studies~\cite{mazilu2012online, koltermann2024gait}. Among evaluated window lengths (2-5 seconds), the 3-second window was optimal, balancing detection latency and data relevance. 
Although intuitively shorter windows (2s) could reduce latency in some cases for larger FoG episodes, our validation revealed that shorter windows contained insufficient contextual information for reliable detection, thus slightly increasing the average detection latency compared to the 3-second window shown in Fig.~\ref{fig:window-size}. Longer windows (3.5s, 4s, 5s) further increased latency without accuracy gains. Therefore, a 3-second window was selected for optimal performance.
\begin{figure*}[t]
\vspace{-5mm}
\centering
\includegraphics[width=0.9\linewidth]{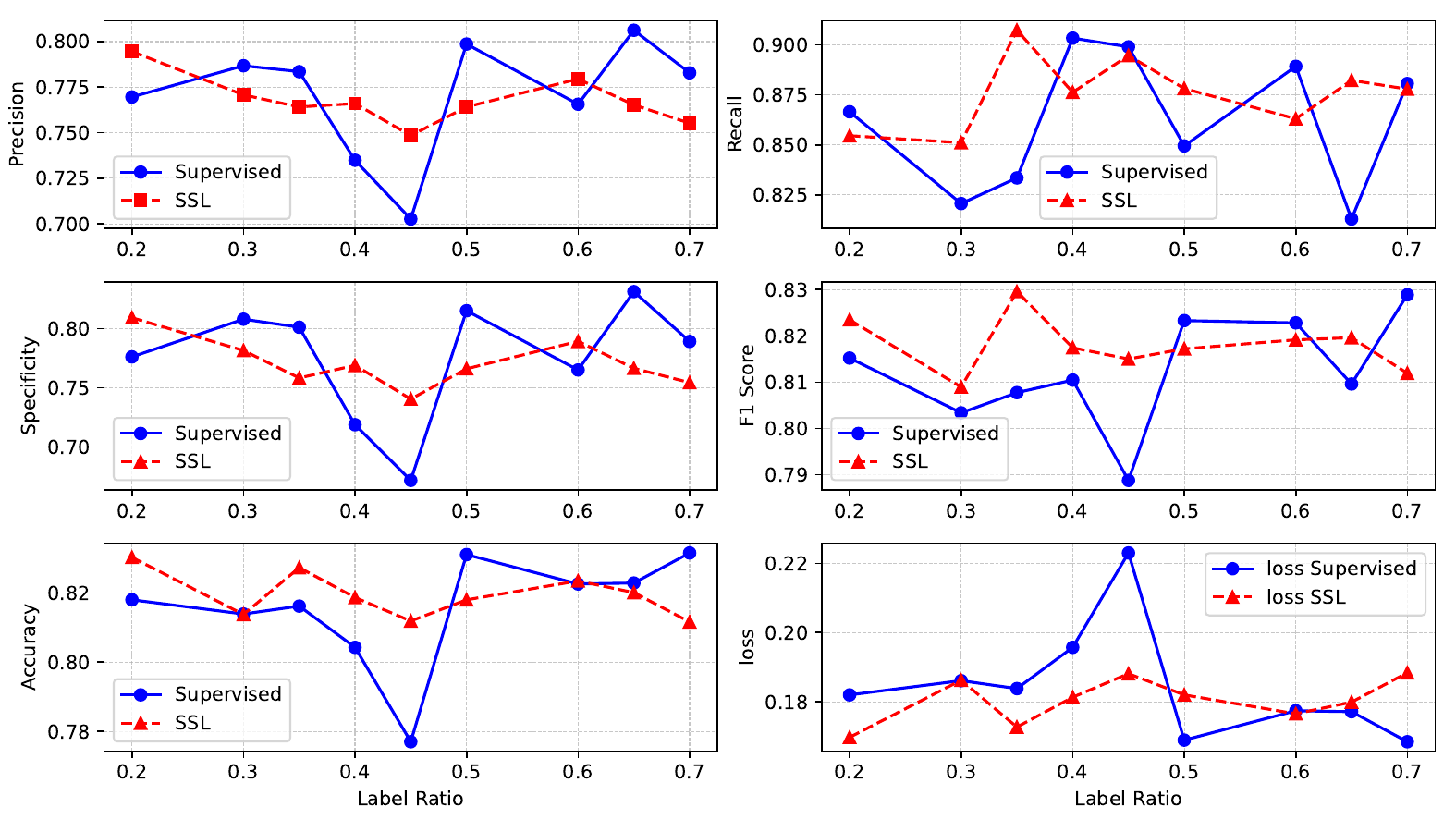}
\vspace{-5mm}
\caption{Performance analysis for different amounts of labeled data.}
\vspace{-4mm}
\label{fig:label-increment}
\end{figure*}
\subsection{Effect of Labels During Pre-training}
Fig.~\ref{fig:label-increment} illustrates how varying the label ratio (on the x-axis) impacts the performance (on the y-axis) of the self-supervised learning (SSL) and supervised models during training.

As the ratio increases from 0.2 to 0.7, 
(meaning more labeled data is available for training), 
both models generally show an upward trend in metrics such as precision, recall, F1-score, and accuracy, which aligns with the expected behavior when more labeled data is available. 
However, a notable observation was the SSL model's superior stability and consistency compared to the supervised counterpart, which displayed sharper performance fluctuations across different label ratios. This steadiness underscores the SSL model's robustness and reduced sensitivity to the availability of labeled data during training. It also highlights the model's capability to leverage pre-training on unlabeled data to learn generalized features transferable to the target task.

Fig.~\ref{fig:label-increment} also shows that SSL model not only competes closely but sometimes outperforms the supervised model, especially with 40-60\% labeled data. This indicates that the SSL model can achieve promising results with fewer labeled instances, making it highly efficient and adaptable for scenarios where obtaining large amounts of labeled data is challenging.

\subsection{Post Processing}
The post-processing analysis evaluated the temporal dynamics between false positives (FPs) and true FoG episodes (TFE) detected by the LIFT-PD framework.


On average, FPs occurred 16 seconds after the previous TFE and 18.5 seconds before the next true episode, suggesting they often arise from residual motor instability or sensor inaccuracies immediately following an actual FoG event.
Our `Pre FP Analysis' shows that FPs were found to occur approximately 14 seconds away from the nearest TFE, indicating the system's high sensitivity to subtle motor pattern changes preceding or following a FoG episode. These false alarms could serve as precursors or warnings of impending FoG events, providing valuable time windows for treatment adjustments.

To address isolated false detections, a majority voting scheme was implemented. If a window's classification differed from its immediate neighbors, it was adjusted to match them, smoothing the detection sequence. This approach improved the detection performance, increasing the detected true FoG episodes from 88\% to 89.8\% and true FoG windows from 83.85\% to 84.64\%.



\section{Limitations}
While LIFT-PD demonstrates promising results and practical advantages for real-world deployment, several limitations remain. First, although the model generalizes well across diverse datasets and patient groups, further validation in larger patient cohorts with varying clinical conditions and disease progression stages, including patients exhibiting tremor-dominant or other atypical Parkinsonian motor symptoms, is needed to confirm robustness and clinical applicability. Specifically, additional assessment is required to determine the model's accuracy in detecting FoG episodes in the presence of confounding symptoms such as resting tremors, dyskinesia, or bradykinesia, which often coexist in individuals with Parkinson's disease. Second, while the system operates effectively with a single accelerometer, incorporating additional sensor modalities such as gyroscopes or physiological sensors (i.e., heart rate, skin conductance) might enhance detection accuracy, particularly for subtle or early-stage FoG events. Third, the Opportunistic Inference Module (OIM), though significantly improving energy efficiency, might occasionally delay activation, potentially missing very brief or subtle FoG events. Fourth, although decision thresholds in OIM were empirically optimized using a small validation subset from the training data, this threshold selection process remains dataset-specific and may not generalize across varying conditions or sensor configurations. We acknowledge that free-living environments introduce high variability, and are currently investigating dataset-agnostic calibration techniques using an independent dataset under acquisition. This future work aims to improve the robustness of threshold selection and ensure consistent performance across diverse deployment settings.\newline
Finally, despite using significantly fewer labeled samples through self-supervised learning, the initial tuning of parameters and training still relies partially on labeled data, and completely unsupervised adaptation remains an open challenge. Future work should address these aspects to further optimize performance, clinical relevance, and adaptability of LIFT-PD.

\section{Discussion}
In this study, we introduced LIFT-PD, a computationally efficient and robust self-supervised learning framework designed specifically for real-time, patient-independent detection of Freezing of Gait (FoG) episodes in patients with Parkinson’s Disease (PD). Our proposed approach addresses key challenges inherent in continuous in-home FoG monitoring, particularly the reliance on large, extensively labeled datasets and substantial power consumption. 

By implementing a novel Differential Hopping Windowing Technique (DHWT), LIFT-PD effectively handles imbalanced data and diverse gait variations, thereby reducing annotation burdens and enhancing clinical scalability. Furthermore, the integration of an opportunistic inference module substantially decreases power consumption by activating the deep learning model exclusively during ambulatory periods. Such selective activation translates to a significant improvement in battery life, making continuous ($>$48-hour) wearable monitoring practical. 

This system requires minimal data preprocessing, utilizing only a single triaxial accelerometer placed comfortably at the waist—an approach traditionally considered suboptimal yet proven highly effective through our self-supervised learning strategy. Clinically, accurate and real-time FoG detection as provided by LIFT-PD enables timely cue delivery (e.g., rhythmic auditory, visual, or vibrotactile stimuli), critical for breaking or preventing episodes without causing habituation~\cite{nieuwboer2008cueing, cosentino2023one}. 

To evaluate the clinical robustness of our model, we stratified performance by disease severity, age, gender, and medication state. These subgroups reflect known sources of variability in Parkinson’s Disease presentation and management. LIFT-PD consistently achieved high F1 scores and specificity across all groups, with particularly strong performance in older adults and those in the On-medication state. The system's ability to robustly detect FoG episodes across varying patient demographics—including different severities, medication states, and age groups—further emphasizes its practical applicability for real-world clinical scenarios, enabling more personalized patient management.

\textcolor{black}{Most mis-detections occur in short, transient FoG episodes (<6s) where the transition dynamics dominate the window and movement magnitude is low. Such cases challenge both temporal continuity and MAM activation sensitivity. Adaptive thresholding and shorter, overlapping inference windows are promising directions to further reduce these errors. While the detection rate for short episodes is lower than for longer episodes, it is important to contextualize this performance within clinical relevance. Short FoG episodes (<6 seconds) typically have less severe impact on patient mobility and fall risk compared to prolonged episodes. However, they may serve as early indicators of medication wearing off or precursors to longer episodes. The average detection latency of 2.42 ± 0.45 seconds for detected short episodes (Table 10) still provides a useful window for preventive cueing interventions.}

\textcolor{black}{
In addition, we compared LIFT-PD against multiple state-of-the-art semi-supervised learning (Semi-SL) approaches, including the published Mikos et al. model~\cite{mikos2017real}, a $\Pi$-model baseline, and a Mean-Teacher (MT) variant, all evaluated using the same data splits and architecture. While these Semi-SL methods leveraged unlabeled data through consistency regularization, they required a larger proportion of labeled samples to achieve stable performance and showed reduced sensitivity in detecting brief FoG events. In contrast, the proposed self-supervised learning (SSL) framework achieved superior recall and F1-scores while using fewer labeled data (~40\%), demonstrating its ability to learn robust, label-agnostic gait representations that generalize effectively across subjects and clinical conditions.
}


\section{Conlusion}
In conclusion, LIFT-PD presents a practical and clinically meaningful solution for real-time detection of Freezing of Gait episodes in Parkinson’s Disease. The system successfully achieves robust, patient-independent monitoring through innovative self-supervised learning and opportunistic inference strategies, reducing dependency on labeled data and significantly improving battery life. Its reliance on a single waist-worn accelerometer ensures patient comfort and adherence, thus enhancing feasibility for continuous at-home use. Furthermore, LIFT-PD demonstrates robust generalization across age, gender, disease severity, and medication states, and its outputs show significant correlation with clinical assessments (e.g., NFOGQ and UPDRS-III Off scores), reinforcing its translational potential. Ultimately, by delivering accurate and timely FoG detection and supporting targeted, personalized cue-based interventions, LIFT-PD significantly advances in-home monitoring capabilities, improves PD symptom management, and contributes positively to patients’ quality of life.

\section*{Data \& Code Availability}
This study utilized the publicly available tDCS FoG dataset. The dataset comprises wearable 3D lower-back sensor recordings annotated for freezing of gait \textcolor{red}{(FoG)} episodes in individuals with Parkinson’s disease. Ground truth labels for three event classes—Start Hesitation, Turn, and Walking—were derived through expert frame-by-frame video annotation. Access to the dataset is available via the competition website: \href{https://kaggle.com/competitions/tlvmc-parkinsons-freezing-gait-prediction}{\textcolor{blue}{https://kaggle.com/competitions/tlvmc-parkinsons-freezing-gait-prediction}}

The codebase for the LIFT-PD framework, including model implementation, training scripts, and evaluation pipelines, is publicly available at: \href{https://github.com/shovito66/LIFT-PD}{\textcolor{blue}{https://github.com/shovito66/LIFT-PD}}

\bibliographystyle{ACM-Reference-Format}
\bibliography{sample}

\end{document}